\documentclass{article}

\usepackage{arxiv}

\usepackage[utf8]{inputenc} 
\usepackage[T1]{fontenc}    
\usepackage{hyperref}       
\usepackage{url}            
\usepackage{booktabs}       
\usepackage{amsfonts}       
\usepackage{nicefrac}       
\usepackage{microtype}      
\usepackage{lipsum}		
\usepackage{amsthm}
\usepackage{graphicx}
\usepackage{doi}
\usepackage{physics}
\usepackage{amssymb}
\usepackage{pifont}

\newcommand{\pd}[2]{\frac{\partial #1}{\partial #2}}

\newcommand{\bs}{\boldsymbol}
\newcommand{\mbf}{\mathbf}

\usepackage{setspace}
\title{Nonlinear elliptic homogenization with parametric Deep Ritz method}

\date{} 					

\author{
Conor Rowan \\
Smead Aerospace Engineering Sciences\\
University of Colorado Boulder\\
3775 Discovery Drive \\
Boulder, CO 80309 \\
\texttt{conor.rowan@colorado.edu} \\
\And
Alireza Doostan \\
Smead Aerospace Engineering Sciences\\
University of Colorado Boulder\\
3775 Discovery Drive \\
Boulder, CO 80309 \\
\texttt{alireza.doostan@colorado.edu} \\
\And
Kurt Maute \\
Smead Aerospace Engineering Sciences\\
University of Colorado Boulder\\
3775 Discovery Drive \\
Boulder, CO 80309 \\
\texttt{kurt.maute@colorado.edu} \\
}

\renewcommand{\headeright}{}
\renewcommand{\undertitle}{}
\renewcommand{\shorttitle}{}

\hypersetup{
pdftitle={A template for the arxiv style},
pdfsubject={q-bio.NC, q-bio.QM},
pdfauthor={David S.~Hippocampus, Elias D.~Striatum},
pdfkeywords={First keyword, Second keyword, More},
}

\begin{document}
\maketitle

\begin{abstract}
Elliptic homogenization is used to determine coarse-grained properties of materials with features on small scales for heat transfer and elasticity. When these small-scale features have rapid, periodic fluctuations, the temperature or displacement field corresponding to a ``homogenized'' constitutive relation closely resembles the true solution based on the heterogeneous material. This homogenized behavior of the material is computed from a cell problem, where a cell is defined to be one period of the fluctuating material. In the context of linear elliptic partial differential equations, the homogenized constitutive relation is defined simply by a constant coefficient tensor, but for nonlinear problems, the homogenized response depends on the macroscopic state and/or its gradient, thus requiring solutions to parametric cell problems. Furthermore, when computing a numerical solution with the homogenized constitutive relation, it is useful to have a differentiable representation of the solution to the cell problem, as derivatives of the homogenized constitutive relation are required in Newton iterations for the macroscopic state field. In this work, we use the Deep Ritz method to solve the parametric cell problems that arise from nonlinear homogenization. First, we exploit the variational structure of the cell problem, then we discretize the dependence of the cell response on both space and the macroscopic state with a neural network. Enforcing boundary conditions on the cell response strongly with a choice of network architecture, we next use the parametric Deep Ritz method to simultaneously solve the cell problem over a range of macroscopic states and gradients. At the level of the cell problem, we show that this method is accurate, efficient, and offers a continuous and differentiable representation of the cell response over the macroscopic state and gradient. We then show that our parametric representation of the cell response expedites macroscale solutions when compared to a traditional $\text{FE}^2$ scheme, thus facilitating numerical solutions to nonlinear multiscale problems.
\end{abstract}



\section{Introduction}

\paragraph{} Statistical mechanics has furnished an impressive array of techniques to extract coarse-grained laws from systems comprising a very large number of interacting parts. However, in the context of real-world engineering problems, it is often intractable to extract coarse-grained constitutive relations from a micro-scale model of the system. Furthermore, engineering materials often have heterogeneities not only on the microscale (atoms, molecules, etc.), but also mesoscale heterogeneity. This is certainly the case in composite materials, where the phase of the material varies on the millimeter and centimeter scale, perhaps between a matrix and an inclusion phase. Phenomenological thermodynamic potentials (entropy, energy, and Legendre transforms thereof) are often sufficient for constitutive modeling of individual phases, and coarse-graining the mesoscale heterogeneity is the problem of practical interest \cite{callen_thermodynamics_1991, holzapfel_nonlinear_2000}. In engineering, the problem of extracting macroscopic material properties from mesoscale heterogeneity is called ``homogenization.'' Note that while material heterogeneity is typically on the mesoscale, these features of the material are referred to as ``microstructure'' in the literature. We adopt this terminology going forward.

\paragraph{} In this work, we focus on elliptic homogenization of materials with periodic fluctuations, with applications in elasticity, heat transfer, and mass transport problems. The theory of elliptic homogenization for linear partial differential equations (PDEs) dates back many decades \cite{bensoussan_asymptotic_1979, bakhvalov_concept_1989}, but has seen many modern treatments, including extensions to nonlinear problems \cite{geers_computational_2010}. The basic idea of elliptic homogenization is to obtain the macroscopic constitutive response by passing displacement, temperature, or concentration gradients through a ``cell'' of the periodically fluctuating material, and then computing the corresponding macroscopic flux by averaging the cell response. A cell is defined to be one period of the fluctuation in the material and is thus the basic building block of the heterogeneous material. For non-periodic or random microstructures, effective properties are instead commonly estimated using a representative volume element (RVE), whose size must be sufficiently large to capture the relevant statistics of the microstructure \cite{kanit_determination_2003, khoromskaia_numerical_2019}. Because we restrict attention to periodic homogenization, we use the terminology of cells of material microstructure. The cell-based approach has been thoroughly explored in the context of the finite element method, where the cell can be meshed, and the cell problem solved numerically, which allows for the computation of effective macroscopic properties \cite{guedes_preprocessing_1990, hollister_comparison_1992, gusev_representative_1997}. Computing macroscopic properties in this way obviates building a finite element mesh to resolve the microstructural fluctuations within the macroscopic domain, which leads to expensive simulations \cite{bishop_direct_2015}. We note that there are a variety of methods to solve multiscale problems which do not rely on the cell response, e.g., the variational multiscale method \cite{hughes_multiscale_1995}, the multiscale finite element method \cite{hou_multiscale_1997}, and the localized orthogonal decomposition \cite{malqvist_localization_2014}. Because our focus is on solving cell problems, we do not consider methods of this sort going forward.

\paragraph{} For linear homogenization problems, the homogenized constitutive relation is also linear, and can thus be computed with the response of the cell to unit gradients. For nonlinear homogenization, the homogenized constitutive response depends nonlinearly on the macroscopic state and its gradient, requiring a tighter coupling between the macroscopic domain and the cell problem, as the cell response needs to be re-computed as the macroscopic state evolves with Newton iterations \cite{geers_multi-scale_2010, smit_prediction_1998, feyel_fe2_2000}. One popular method for passing information across scales is the $\text{FE}^2$ approach, which solves a finite element problem for the cell response at each macroscopic integration point, and computes the constitutive response accordingly \cite{feyel_fe2_2000}. $\text{FE}^2$ has been used successfully with the finite element method in the context of thermoelasticity \cite{ozdemir_fe2_2008}, piezoelectric materials \cite{khalaquzzaman_computational_2010}, and multiscale fracture mechanics \cite{coenen_multi-scale_2012}. Newton iterations at the macroscale additionally require derivatives of the homogenized constitutive relation with respect to its macroscopic arguments. In conventional $\text{FE}^2$, evaluation of this consistent tangent requires linearization or sensitivity analysis of the cell problem \cite{miehe_computational_2002, miehe_computational_2003, temizer_computation_2008}.


\paragraph{} In the last decade, scientific machine learning (SciML) has offered many new perspectives on classical problems in computational science, as well as introduced several fundamentally new tools. From 2017 to 2019, physics-informed neural networks (PINNs) established a new paradigm for scientific computing, where data and physical knowledge could be seamlessly integrated \cite{raissi_physics-informed_2019, sirignano_dgm_2018}. With PINNs, neural networks discretize the PDE solution field, and the loss function used to train the network can incorporate both the governing differential equation and measurement data on the system of interest. This framework streamlines numerical solutions to complex PDEs, does not require a mesh, and requires minimal modification to solve inverse problems \cite{cai_physics-informed_2021}. Some variations to the basic PINNs framework involve using different loss functions, for example, the variational energy \cite{e_deep_2017, abueidda_deep_2022, liu_deep_2023} or the weak form of the governing equation \cite{khodayi-mehr_varnet_2019}. PINNs have been successfully used in many application areas, such as fracture mechanics \cite{manav_phase-field_2024}, fluid mechanics \cite{jin_nsfnets_2021}, nonlinear elasticity \cite{abueidda_deep_2022}, and heat transfer \cite{cai_physics-informed_2021}. Despite their successes, PINNs rely on knowledge of the governing equation; training can be expensive, and they require re-training if the PDE input is changed. In response to these challenges, a technique called operator learning trains neural networks to predict the PDE solution function over a range of problem inputs entirely from data. This was first explored with the Deep Operator Network \cite{lu_deeponet_2021}, and the basic idea has given rise to several other architectures, such as the Fourier Neural Operator \cite{li_fourier_2021}, the Laplace Neural Operator \cite{cao_lno_2023}, Wavelet Neural Operator \cite{tripura_wavelet_2023}, among others \cite{ingebrand_basis--basis_2025, lanthaler_operator_2023, melchers_neural_2025, li_neural_2020}. Two noteworthy benefits of neural operators are that knowledge of the governing equation is not required in order to construct a predictive model, and they are efficient to evaluate once trained, as the operator acts as a surrogate for solving a PDE, thus avoiding computationally expensive solutions to large systems of equations. That being said, we remark that when it is available, it is possible to include knowledge of the physics to the operator learning process \cite{li_physics-informed_2023}.

\paragraph{} PINNs and neural operators have been studied extensively in many scientific fields, but there are comparatively few works using the techniques of SciML for homogenization. Because cell problems require potentially expensive numerical solutions to determine material properties, and the cell response cannot always be characterized by unit-load solutions, a number of authors have explored using data-driven methods as a surrogate for the cell problem. An early example of using machine learning tactics to accelerate nonlinear homogenization relies on neural networks to approximate the relationship between the deformation and homogenized stress tensor in elastic solids \cite{dana_machine_2021}. In \cite{gupta_accelerated_2023}, a data-driven model is trained to predict the stress field in the cell of a composite material, which allows for efficient computation of macroscopic properties. Similarly, neural network models have been trained to map from macroscopic electric field to effective current vectors \cite{lu_data-driven_2019}. Another approach is to use a neural network to approximate the effective strain energy density as a function of macroscopic variables and microstructural parameters, allowing the homogenized stress to be obtained through differentiation \cite{le_computational_2015}. A particularly simple strategy is to leverage knowledge that linear elliptic equations admit homogenized solutions, and directly learn the homogenized tensor from data. This approach was pursued in \cite{park_physics-informed_2022}, where the authors use data on a multiscale system to approximate the homogenized coefficients by solving an inverse problem with PINNs. Other works have explored the use of Fourier Neural Operators for homogenization of linear elliptic equations \cite{stepanov_neural_2026, nguyen_universal_2026}. Neural surrogate models for the constitutive response of multiscale solids have also been explored in the context of structural design optimization \cite{black_neural_2025}. Neural surrogates are differentiable with respect to the macroscopic state and its gradient, but they do not encode physics, relying instead on large quantities of training data. Furthermore, if the surrogate maps from the macroscopic state and gradient to the homogenized response directly, it avoids the cell problem, which provides important information about the maximum temperature or stress in the body \cite{torisaki_shape_2023}.

\paragraph{} By solving the cell problem numerically with a neural network discretization, other machine learning-based approaches to homogenization rely on physics over data. An early example of this solves the cell problem with a neural network discretization, computes the homogenized material properties from the cell problem, and then solves the homogenized macroscopic equation with a PINN approach \cite{leung_nh-pinn_2022}. As the authors discuss, this allows the neural network to accurately model the macroscopic response of the structure, as neural networks struggle to represent high-frequency solution fields, a phenomenon known as the ``spectral bias'' \cite{rahaman_spectral_2019}. Relying on methods to strongly enforce periodic boundary conditions introduced in \cite{dong_method_2021}, subsequent works have also used neural network discretizations to solve the cell problem, whose solution governs the macroscopic material response. In \cite{soyarslan_physics-informed_2024}, the authors use exact boundary condition enforcement to homogenize three-dimensional microstructures for linear elliptic problems. Similar methods were explored in the multi-physics setting, where the homogenized electromechanical properties of periodic composites were computed using neural network discretizations of the cell problem \cite{chen_deep_2024}. Again investigating periodic composites, another study uses domain decomposition to split the cell problem into matrix and inclusion sub-domains, with neural networks discretizing the solution field in each phase \cite{chen_hybrid_2025}. In \cite{zhao_physically_2025}, the authors use neural networks to discretize the solution to the cell problem, but the neural network architecture and loss function are modified to accommodate high-frequency responses and sharp gradients in the cell. Noting that PINN solutions can fail in the presence of discontinuous coefficient fields, the authors in \cite{gaynutdinova_homogenization_2026} introduce a novel formulation of the PINN loss that is more robust to these discontinuities. For a thorough review of machine learning approaches to multiscale problems, see \cite{bishara_state---art_2023}. We note that the above studies focus primarily on linear homogenization, where the cell problem is not influenced by the macroscopic state of the body. An exception to this is given by a recent work, which uses a variational formulation of the cell problem to compute homogenized material properties in the context of hyperelasticity \cite{li_energy-based_2026}. In this work, the authors use the Deep Ritz method to solve for the homogenized response of the cell at given macroscopic parameters. If their approach were to be included in an $\text{FE}^2$ scheme, the network governing the cell response would need to be continually retrained as the macroscopic state evolved with Newton iterations. This motivates a parametric formulation of the cell problem for nonlinear homogenization, where the network discretizes the cell response in both physical space and in the space of macroscopic parameters. A trained network of this sort fully avoids online solves of the cell problem when computing the macroscopic response, thus expediting the solution process. To the best of our knowledge, parametric solution strategies for homogenization have not been studied in the context of neural networks; they have been treated with classical discretizations under the title of proper generalized decomposition (PGD). Generally speaking, PGD is a framework for solving parametric PDEs where the solution is discretized with a low-rank separated representation over physical and parameter spaces \cite{chinesta_proper_2010}. Nonlinear homogenization with PGD also avoids online cell solutions, and has been treated in various works \cite{cremonesi_pgd-based_2013, el_halabi_pgd-based_2016}. In particular, PGD represents the corrector as a sum of products of functions of the cell coordinate, the macroscopic temperature, and its gradient. When the energy density is separable, the integrals involved in generating the PGD factorize into low-dimensional ones, and no product grid is needed. However, for the general non-separable settings, the standard approach is to first approximate relevant quantities, e.g., material properties, in separated form and then contract \cite{zlotnik_effect_2015}; otherwise, quadrature over the coupled variables is required, whose cost grows exponentially with their number. Sampling-based quadrature can also be combined with a separated ansatz, at the cost of quadrature error and, in greedy constructions, of the guarantees established for the exact energy.
%
%
In the present study, we take the closest relative of our proposed method to be \cite{li_energy-based_2026}, where the cell problem is solved with a non-parametric Deep Ritz method.


\paragraph{} In this work, we leverage the ability of neural networks to handle parametric problems to simultaneously solve for the cell's response over a range of macroscopic states and gradients. Our contributions are as follows:

\begin{enumerate}
    \item We formulate the cell problems of nonlinear elliptic homogenization parametrically and variationally, allowing us to solve them with the parametric Deep Ritz method; 
    \item We demonstrate that this leads to accurate and efficient solutions to the cell problem, which are differentiable with respect to macroscopic parameters;
    \item In the context of repeated solutions of the cell problem, we compare against a non-parametric neural network discretization and a traditional $\text{FE}^2$ implementation of a nonlinear multiscale problem, showing that the parametric representation of the cell response furnished by our method dramatically expedites the numerical solution process.
\end{enumerate}

The rest of this paper is organized as follows. In Section 2, we outline our approach to nonlinear homogenization, which does not refer to perturbation theory or asymptotic expansions. In Section 3, we introduce the parametric Deep Ritz method for the cell problems of nonlinear homogenization, showing how periodic boundary conditions can be enforced strongly with neural network discretizations. In Section 4, we provide numerical examples from different problem types in nonlinear heat transfer, showcasing the ability of our method to handle a varying number of macroscopic state parameters. In Section 5, we solve a nonlinear multiscale problem with a non-parametric Deep Ritz method and a traditional $\text{FE}^2$ approach, showing that the parametric representation of the cell response avoids online numerical solutions of the cell problem, thus expediting the solution process in repeated query applications such as homogenization with $\text{FE}^2$. In Section 6, we close with concluding remarks and directions for future work.


\section{Nonlinear homogenization}

\paragraph{} In this work, we focus on homogenization problems from nonlinear heat transfer, though we note that this is not a limitation of our method. For the sake of clarity, we first present the case of standard Fourier-type heat conduction, where the thermal conductivity of the material is temperature dependent. When the conductivity has rapid periodic fluctuations in space, this leads to a special nonlinear homogenization problem where the cell problem is linear but parameterized by the macroscopic temperature. We then consider the case of $p$-Laplace heat conduction, where the cell problem becomes nonlinear and is parameterized by both the macroscopic temperature and temperature gradient. This section is a review of homogenization theory and is used to motivate the variational formulation of the cell problem.

\subsection{Multiscale Fourier-type nonlinear heat transfer}

\paragraph{} A basic motivation for elliptic homogenization comes from the following problem:
\begin{equation}\label{hom_motivation}
    \pd{}{x}\qty( \kappa(x/\eta) \pd{u_0}{x}) = 0, \quad u_0(0)=0, \quad \kappa(1/\eta)\pd{u_0}{x}(1)=1,
\end{equation}
\noindent where $u_0(x)$ is the temperature, $x$ is the spatial coordinate, and the material property $\kappa(x/\eta)$ is periodic in its argument, with the parameter $\eta$ controlling the ``scale'' of the problem. In particular, as $\eta \rightarrow 0$, the material fluctuates at increasingly high frequencies. An example material of this sort is given by 
\begin{equation*}
    \kappa(x/\eta) = 1 + \frac{1}{2}\sin(x/\eta).
\end{equation*}

By solving Eq. \eqref{hom_motivation} with different values of $\eta$, one sees that as $\eta$ approaches zero, the state ceases to oscillate with the material and begins to behave as if the material were constant. The goal of linear elliptic homogenization is to find the value of this fictitious constant material property as a function of the multiscale material $\kappa(x/\eta)$. This idea of the fluctuations in the solution decaying away with small $\eta$ is the basic motivation for both linear and nonlinear homogenization.

\begin{figure}[hbt!]
\centering
\includegraphics[width=0.75\textwidth]{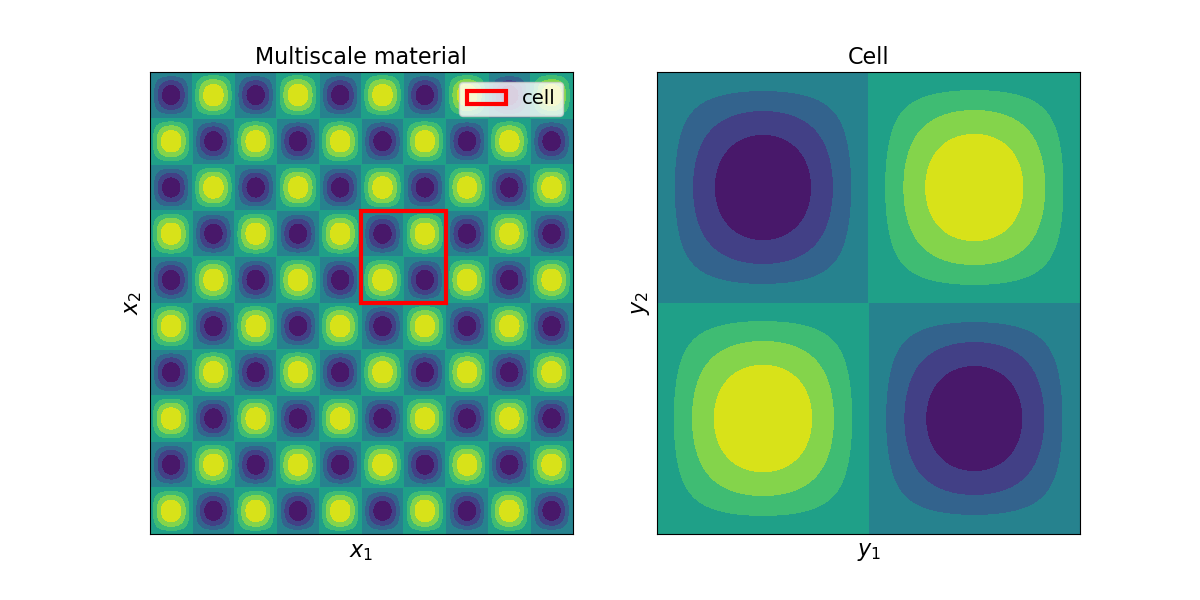}
\caption{Extracting a cell of microstructure from a macroscopic domain with a periodic material property. The material property in the cell is indexed by $\mbf y$, and represents one period of the fluctuation.}
\label{cell_definition}
\end{figure}

\paragraph{} We consider the general case of nonlinear Fourier-type multiscale heat transfer in any number of spatial dimensions, though our arguments will be presented in two spatial dimensions for ease of visualization. It is conceptually straightforward to generalize these arguments to three dimensions. By nonlinear Fourier-type heat transfer, we mean that the heat flux is linear in the temperature gradient, with a coefficient of proportionality that is temperature dependent. We take the multiscale, temperature-dependent material property to be isotropic. The governing equation is thus 
\begin{equation}\label{multiscale_problem}
    \nabla_x \cdot ( \kappa( u_0, \mbf x/\eta) \nabla_x u_0(\mbf x) ) + f(\mbf x) = 0 , \quad \mbf x \in \Omega_x,
\end{equation}
\noindent where $u_0$ is the temperature, $f(\mbf x)$ is a source term, and $\Omega_x$ defines the geometry of the body under consideration. Note that, by assumption, the conductivity is periodic in its spatial argument, but not necessarily in its temperature dependence. The goal of the nonlinear homogenization problem is to rid the conductivity of its fast fluctuating spatial component. We will see that this can be accomplished, leading to a homogenized conductivity with only temperature dependence, which we call $\bs{\hat \kappa }(u_0)$.\footnote{We assume throughout that the constitutive relation has no ``slow'' spatial dependence. In other words, it could be that the conductivity has the form $\kappa(u_0,\mbf x , \mbf x /\eta)$, where there is a periodic fluctuation given by $\mbf x / \eta$ and a slower dependence given by the $\mbf x$ argument. Physically, this could be a functionally graded material with multiscale properties. In these cases, the task of homogenization is to rid the conductivity only of the dependence on the fast variable, while the slow dependence remains.} To do this, we introduce the idea of a cell, which contains one period of the fluctuating material. See Figure \ref{cell_definition} for the definition of a cell with an example multiscale material. We index positions in the cell with a new spatial coordinate $\mbf y$ and call the cell domain $\Omega_y$. Note that the cell domain is taken to be square, but its size is not important, so we say generically that $\Omega_y = [0,L]^2$. We denote the material property distribution in the cell as $\kappa( u_0 , \mbf y)$. 

\begin{figure}[hbt!]
\centering
\includegraphics[width=0.8\textwidth]{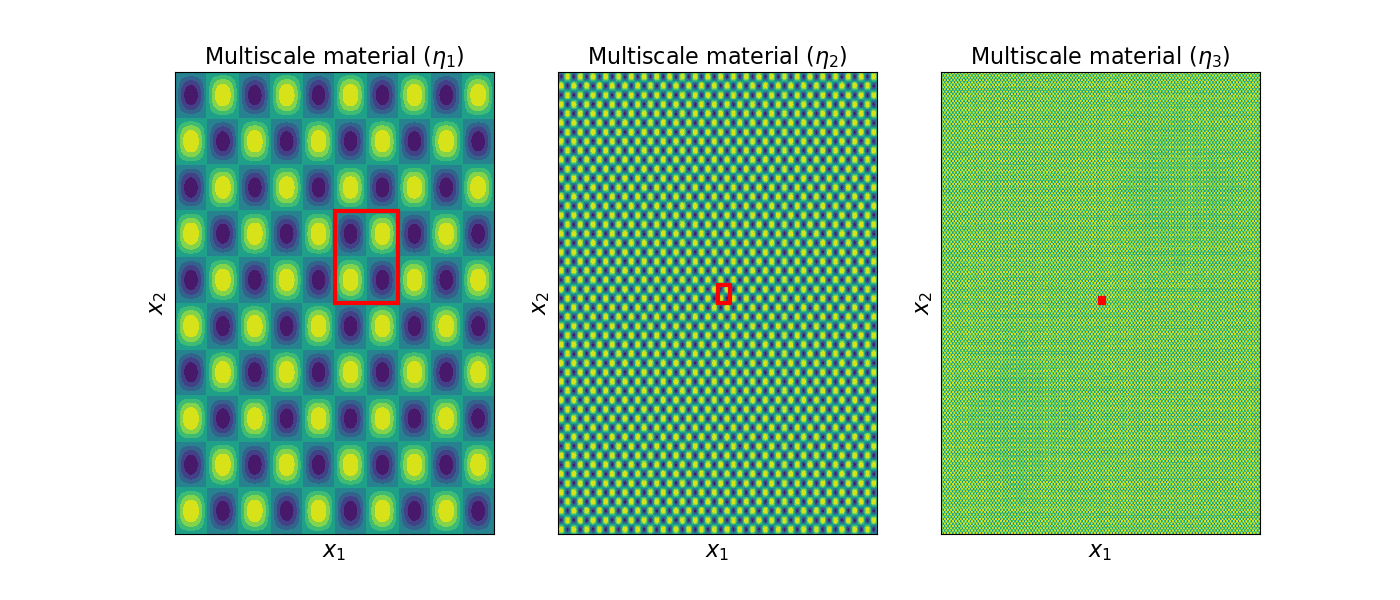}
\caption{As the scale parameter $\eta$ gets small, the fluctuations in the material become higher frequency and the cell appears to occupy a single point in the macroscopic domain. The idea of elliptic homogenization is to treat the constitutive relation at a point to be determined by the response of the cell to the temperature gradient at that point. }
\label{limiting}
\end{figure}

\paragraph{} With elliptic homogenization, we argue that the cell determines the effective constitutive response of the multiscale body. In the limit that $\eta \rightarrow 0$, the cell occupies a single point in space $\mbf x^*$, and the variation of the temperature field across the cell is accurately approximated with a linearization $u_0(\mbf x) \approx u_0(\mbf x^*) + \nabla_x u_0(\mbf  x^*) \cdot (\mbf x - \mbf x^*)$. The homogenized response of the material is then obtained by finding the total heat flux through the cell arising from this imposed temperature distribution. We will define what is meant by ``total heat flux'' shortly. To compute the cell's response to the macroscopic applied temperature field---macroscopic in that it originates from the domain $\Omega_x$---we set up a boundary value problem for the temperature field in the cell. This is given by 
\begin{equation}
    \nabla_y \cdot (  \kappa( u_0(\mbf x^*) , \mbf y) \nabla_y u(\mbf y) ) = 0, \quad \mbf y \in \Omega_y,
\end{equation}
\noindent where $u(\mbf y)$ is the temperature in the cell and $\kappa( u_0(\mbf x^*), \mbf y)$ gives the conductivity of the cell, whose temperature dependence is fixed by the macroscopic temperature at the given point $\mbf x^*$. Now, using the fact that the response of the cell is driven by the macroscopic temperature gradient, we can write
\begin{equation}\label{ansatz}
    u(\mbf y) = u_0(\mbf x^*) + \begin{bmatrix}
        \pd{u_0}{x_1}(\mbf x^*) \\ \pd{u_0}{x_2}(\mbf x^*)
    \end{bmatrix} \cdot \begin{bmatrix}
         y_1  \\
         y_2 
    \end{bmatrix}+ \chi(\mbf y),
\end{equation}
\noindent where the first term sets the macroscopic reference temperature in the cell, the second term applies the given temperature gradient across the cell, and the third term is what we call the ``corrector,'' as it accounts for the fact that the temperature in the cell need not be linear. Plugging the temperature distribution into the cell problem, we obtain
\begin{equation}\label{corrector_bvp}
    \nabla_y \cdot( \kappa( u_0 , \mbf y) \nabla_y \chi(u_0, \nabla_x u_0 , \mbf y) ) + \nabla_y \kappa( u_0 , \mbf y) \cdot \nabla_x u_0 = 0,
\end{equation}
\noindent where we drop the dependence of the macroscopic quantities on the position $\mbf x^*$ for readability. Note that this problem is forced by the macroscopic temperature gradient $\nabla_x u_0$, and is linear in this quantity. We can thus write the corrector as $\chi(u_0,\nabla_x u_0 , \mbf y) = \nabla_x u_0 \cdot \bs \chi(u_0,\mbf y)$, where the components $\chi_i(u_0,\mbf y)$ solve for the corrector under a unit macroscopic temperature gradient $\partial u_0/ \partial x_i$ at a given macroscopic temperature $u_0$:
\begin{equation}\label{cell_problem_unit}
    \nabla_y \cdot( \kappa(u_0,\mbf y) \nabla_y \chi_i(u_0, \mbf y)) + \pd{\kappa(u_0, \mbf y)}{y_i} = 0.
\end{equation}

We remark that this problem has a variational structure, meaning that each corrector can be written as the minimizer of a suitably defined energy functional for a given macroscopic temperature $u_0$. Because the variational problems are independent for each corrector, we can also combine them into one vector-valued energy functional for the unit corrector vector:
\begin{equation*}
    \underset{\bs \chi (u_0 , \mbf y)}{\text{argmin }} \int \frac{1}{2} \kappa( u_0 , \mbf y) \qty( \pd{\chi_i}{y_j} \pd{\chi_i}{y_j}) - \pd{\kappa}{y_i} \chi_i d\mbf y.
\end{equation*}

Integrating by parts the gradient of the conductivity in the second term, we define the variational energy functional whose minimum governs the unit correctors as
\begin{equation}\label{variational_boundary}
    \Pi( \bs \chi( u_0 , \mbf y) ) = \int \frac{1}{2}\kappa(u_0, \mbf y) \nabla_y \bs \chi : \nabla_y \bs \chi + \kappa(u_0, \mbf y) \nabla_y \cdot \bs \chi d\mbf y - \int \kappa(u_0, \mbf y) \bs \chi \cdot \mbf n dS.
\end{equation}

\paragraph{} So far, nothing has been said about the boundary conditions on the correctors $\bs \chi$, and thus solutions to the cell problem of Eq. \eqref{corrector_bvp} are non-unique. As is standard in the homogenization literature, we take the boundary conditions on the corrector to be periodic \cite{pinho-da-cruz_asymptotic_2009}. The unit responses, packaged in the vector field $\bs \chi$, thus inherit the periodicity of the corrector field. See Appendix A for a justification of periodic boundary conditions. The homogenized response of the cell is given by the average flux through cross-sections aligned with coordinate directions, which, because of periodicity, turns out to be equivalent to a volume average of the flux. This allows us to write the homogenized constitutive relation as
\begin{equation}\label{homogenized}
     q_i( u_0, \nabla_x u_0) = -\qty[\frac{1}{L^2} \int  \kappa( u_0 , \mbf y)  \qty( \delta_{ij} + \pd{\chi_j(u_0,\mbf y)}{y_i})  d \mbf y] \pd{u_0}{x_j} := -\hat \kappa_{ij}(u_0) \pd{u_0}{x_j},
\end{equation}
\noindent where $\bs{\hat \kappa}(u_0)$ is the homogenized constitutive tensor. Note that even with periodic boundary conditions, the corrector is not uniquely defined, as it can be shifted by a constant and still satisfy the governing equation. However, this rigid body mode does not affect the computed constitutive response, as only the gradients of the corrector appear, so we enforce that $\int \bs \chi d\mbf y = 0$. See Appendix B for a discussion of properties that the thermal constitutive relation must obey, which are enforced automatically by Eq. \eqref{homogenized}. We take the automatic enforcement of constitutive properties to be an advantage of going through the cell problem, whereas purely data-driven methods either rely on implicit enforcement of constitutive properties through the training data, or require significant fine-tuning at the level of the model's architecture to enforce them \cite{dana_machine_2021,park_physics-informed_2022, linden_neural_2023, klein_finite_2022}.

\paragraph{} Noting that $u_0$ is simply the macroscopic temperature distribution at the given spatial point, and referring back to the multiscale problem of Eq. \eqref{multiscale_problem}, our homogenized nonlinear Fourier-type heat conduction problem is thus
\begin{equation}\label{homogenized_problem}
    \nabla_x \cdot( \bs{\hat \kappa}(u_0) \nabla_x u_0(\mbf x)) + f(\mbf x) =0, \quad \mbf x \in \Omega_x,
\end{equation}
\noindent where the homogenized conductivity $\bs{\hat{\kappa}}$ has eliminated the spatial dependence of the conductivity field, and may no longer represent an isotropic material. When numerically solving Eq. \eqref{homogenized_problem}, it is necessary to know the homogenized tensor as a continuous function of the temperature. As seen in Eq. \eqref{homogenized}, it is thus necessary to know the corrector components $\bs \chi$ as a function of $u$. Furthermore, when performing Newton iterations, it is necessary to know the derivative $\partial \bs{\hat \kappa}/ \partial u_0$. Thus, providing continuous and differentiable dependence of the corrector on the macroscopic temperature is a desideratum of a numerical method used to solve the cell problem.

\paragraph{} Note that the periodic boundary conditions on the corrector cancel the boundary term in the variational energy of Eq. \eqref{variational_boundary}. Removing the boundary term and noting that shifting by a constant does not change the minimizer of the energy, this expression can be rewritten as 
\begin{equation}\label{varf}
    \Pi(\bs \chi(u_0 , \mbf y)) = \int \frac{1}{2}\kappa(u_0, \mbf y) \qty( \delta_{ij} + \pd{\chi_i}{y_j}) \qty(\delta_{ij} + \pd{\chi_i}{y_j}) d\mbf y. 
\end{equation}

In the following subsection, we derive the homogenized constitutive relation for $p$-Laplace-type heat conduction and the variational problem for the corrector. In this case, the cell problem is nonlinear in the macroscopic temperature gradient, and thus, there is no unit response structure.

\subsection{Multiscale $p$-Laplace heat transfer}

\paragraph{} The $p$-Laplace equation represents a nonlinear extension of heat conduction in which the conductivity of the material depends on the temperature gradient \cite{lindqvist_note_1993}. Again, taking the material property to be temperature-dependent, the $p$-Laplace cell problem is given by 
\begin{equation*}
    \nabla_y \cdot (\kappa(u_0, \mbf y) |\nabla_y u|^{p-2} \nabla_y u) = 0, \quad \mbf y \in \Omega_y,
\end{equation*}
\noindent where the parameter $p>1$ controls the nonlinearity, and we recover the usual Fourier heat conduction with $p=2$. With the same temperature ansatz as Eq. \eqref{ansatz}, the corrector is again forced by the macroscopic temperature gradient, but it is no longer linear in this quantity:
\begin{equation*}
    \nabla_y \cdot( \kappa(u_0 ,\mbf y) | \nabla_x u_0 + \nabla_y \chi |^{p-2} (\nabla_x u_0 + \nabla_y \chi)) = 0, \quad \mbf y \in \Omega_y.
\end{equation*}

Because the cell problem is nonlinear, we cannot decompose the corrector $\chi$ into a linear combination of unit responses. This means the cell problem is parameterized by both the macroscopic temperature $u_0$ and the macroscopic temperature gradient $\nabla_x u_0$. Note that the $p$-Laplace cell problem can be formulated variationally with the following energy functional:
\begin{equation}\label{varp}
    \Pi( \chi(u_0, \nabla_x u_0,\mbf y)) = \int \frac{1}{p}\kappa(u_0, \mbf y)| \nabla_x u_0 + \nabla_y \chi |^{p} d\mbf y.
\end{equation}

As before, periodic boundary conditions on the corrector ensure that fluxes through the cell are well-defined in our intended sense (integrated heat flux independent of position of cross-section), as this is a consequence only of the divergence-free structure of the heat flux. When the corrector solves the cell problem, the homogenized heat flux is
\begin{equation}\label{pflux}
    \mbf q( u_0, \nabla_x u_0) = -\frac{1}{L^2} \int \kappa(u_0 , \mbf y)| \nabla_x u_0 + \nabla_y \chi|^{p-2}( \nabla_x u_0 + \nabla_y \chi ) d\mbf y.
\end{equation}

\section{Parametric Deep Ritz method}

\paragraph{} We turn to the parametric Deep Ritz method to compute the corrector as a continuous function of the parameters of the macroscopic state. Depending on whether the cell problem is linear, we have two different kinds of variational problems. In Eq. \eqref{varf}, a variational problem governs a corrector vector, which packages the cell response from applied unit temperature gradients. The cell problem is thus parameterized only by the macroscopic temperature $u_0$. On the other hand, the $p$-Laplace cell problem does not have a linear structure, which means that the variational problem of Eq. \eqref{varp} is parameterized both by the macroscopic temperature $u_0$ and the temperature gradient $\nabla_x u_0$. To outline the parametric Deep Ritz method for a general variational problem arising from nonlinear homogenization, we call the parameters of the macroscopic state $\mathcal M$, and the quantity of interest $\Gamma$, which is understood to be either a vector of unit responses, or the scalar response to a given temperature and temperature gradient. With the Deep Ritz method, we take the quantity of interest to be discretized by a neural network with trainable parameters $\bs \theta$, thus $\Gamma = \Gamma(\mathcal M , \mbf y; \bs\theta)$. The parametric Deep Ritz method determines the quantity of interest as a continuous and differentiable function of the cell coordinate and macroscopic state parameters with
\begin{equation*}
    \underset{\bs \theta}{\text{argmin }} \int \Big[ \Pi\Big( \Gamma( \mathcal M , \mbf y; \bs \theta) \Big)  + \lambda \Big \lVert \Big \langle \Gamma( \mathcal M , \mbf y; \bs \theta) \Big\rangle \Big \rVert^2 \Big]d\mathcal M,
\end{equation*}
\noindent where $\Pi( \Gamma , \mathcal M)$ is the variational energy whose minimum governs the cell response at a given set of parameters $\mathcal M$ (Eqs. \eqref{varf} and \eqref{varp}, for example). The notation $\langle \Gamma \rangle$ indicates the spatial average, where this term is introduced as a penalty to remove the rigid body mode from the corrector, and $\lambda$ is a penalty parameter. While the mean of the corrector does not influence the computed macroscopic response, it is necessary when using a variational formulation of the cell problem, as depending on the formulation of the energy objective (e.g., whether or not integration by parts is carried out), the energy can be made arbitrarily small with constant shifts in the corrector. The integration over the macroscopic parameters is carried out over some finite region of interest. Note that the corrector must have periodic boundary conditions. Per \cite{dong_method_2021}, this can be enforced strongly by transforming the relevant inputs to the neural network to be periodic. Going forward, we set the size of the cell problem to be $L=1$. The input layer to the neural network representation of the corrector is thus 
\begin{equation}\label{periodic_embedding}
    \begin{bmatrix}
        y_1 \\ \vdots \\ y_D \\ \mathcal M
    \end{bmatrix} \rightarrow \begin{bmatrix}
        \sin(2\pi y_1) \\
        \cos(2\pi y_1) \\
        \vdots \\
        \sin(2\pi y_D) \\
        \cos( 2\pi y_D ) \\
        \mathcal M
        
    \end{bmatrix} ,
\end{equation}
\noindent where $D$ is the number of spatial dimensions. Cell problems arising from nonlinear homogenization are especially well-suited for neural network discretizations because 1) the geometry is naturally square, 2) boundary conditions can be enforced strongly, 3) the problem is variational and thus already an optimization problem, 4) the solution field is not high-frequency, as only one cell is under consideration (thus avoiding concerns about spectral bias), and 5) the continuous dependence of the solution on one or more parameters is required. However, to the best of our knowledge, the parametric Deep Ritz method has not been used in the context of nonlinear homogenization. Note that if the parametric cell problem is solved with classical numerical methods, it is necessary either to extend the discretization into parameter space or to interpolate cell solutions at discrete parameter settings \cite{yamanaka_surrogate_2023}. An additional benefit of the neural network discretization in the case of linear parametric cell problems is that it allows parameters to be shared in the network's representation of the $D$ corrector components, whereas a traditional numerical method must discretize each separately. In the following section, we showcase the accuracy and efficiency of the proposed method for solving parametric cell problems of nonlinear homogenization in one and two spatial dimensions. Subsequently, we show the efficiency gains of having a parametric representation of the corrector field by comparing with a non-parametric Deep Ritz formulation and a traditional $\text{FE}^2$ scheme.


\section{Numerical examples}

\paragraph{} In this section, numerical examples are provided to illustrate the accuracy of the novel parametric Deep Ritz formulation for solving the cell problems of nonlinear homogenization. In the next section, we illustrate the computational efficiency gains in constructing a parametric cell response for the multi-query-type applications like the $\text{FE}^2$ method.

\subsection{One-dimensional Fourier-type heat conduction}

\paragraph{} For our first example, we consider homogenization of the following nonlinear boundary value problem:
\begin{equation}\label{1d_ms}
\begin{aligned}
    \pd{}{x}\qty(\kappa(u_0, x/\eta) \pd{u_0}{x}) + f_0 =0, \quad x\in[0,1],\\
    u_0(0)=0, \quad  \pd{u_0}{x}(1)=0, \\
    \kappa(u_0,x/\eta) = 1 + 0.9 \tanh(2.5u_0 \sin( 2\pi x /\eta)), \\
    \kappa(u_0,y) = 1 + 0.9 \tanh(2.5u_0 \sin( 2\pi y)).
\end{aligned}
\end{equation}

\begin{figure}[hbt!]
\centering
\includegraphics[width=0.85\textwidth]{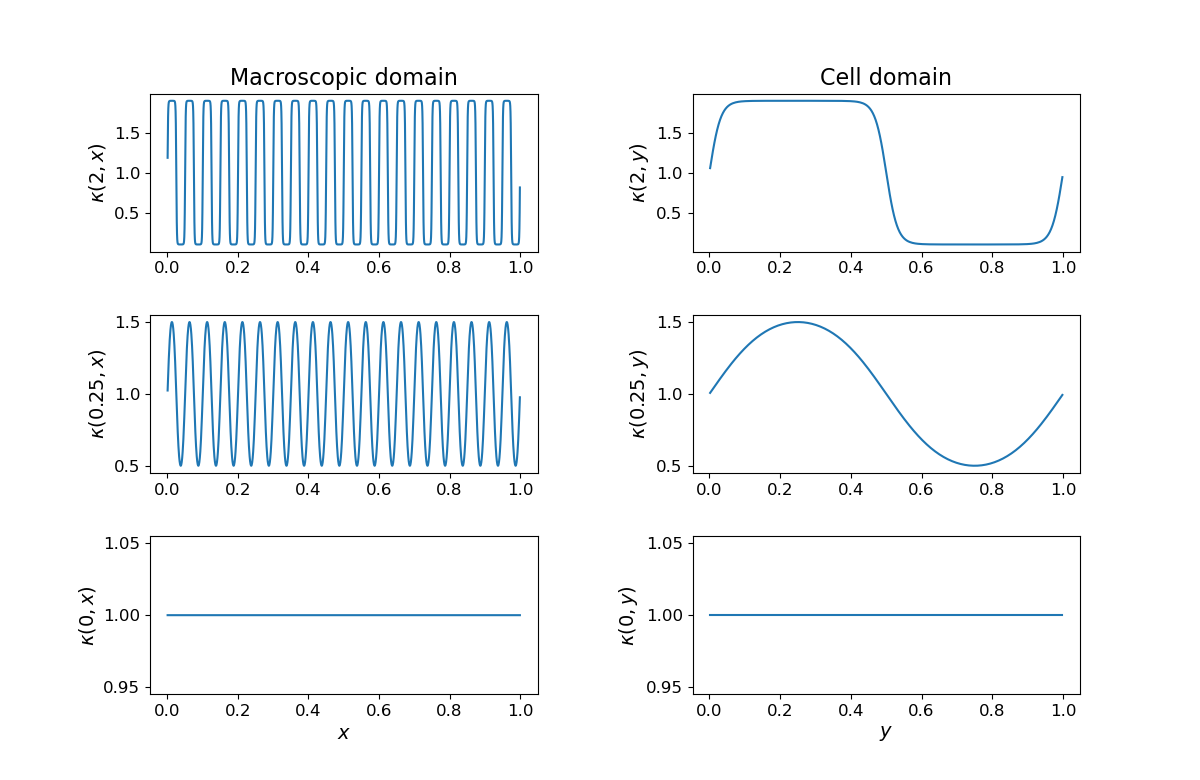}
\caption{Visualizing the temperature-dependent multiscale conductivity and the cell material properties at different temperatures. Both the amplitude and the shape of the fluctuation in the material property change with the macroscopic temperature.}
\label{problem1setup}
\end{figure}

See Figure \ref{problem1setup} to visualize the material field in the macroscopic and cell domains at $\eta=1/10$ for three different temperatures. This is an interesting model problem, as the temperature does not simply scale up and down the conductivity of the fluctuating material in the cell; rather, it changes the shape of the fluctuations. 

\paragraph{} The first thing we do is solve the multiscale problem in a brute force way with a classical numerical method to generate a benchmark against which to compare the homogenized solution. We set $f_0=1$ and discretize the temperature as
\begin{equation*}
    u_0(x) = \sum_{i=1}^N u_i h_i(x) = \sum_{i=1}^N u_i \sin\qty( \frac{2i-1}{2} \pi x),
\end{equation*}

\begin{figure}[hbt!]
\centering
\includegraphics[width=0.85\textwidth]{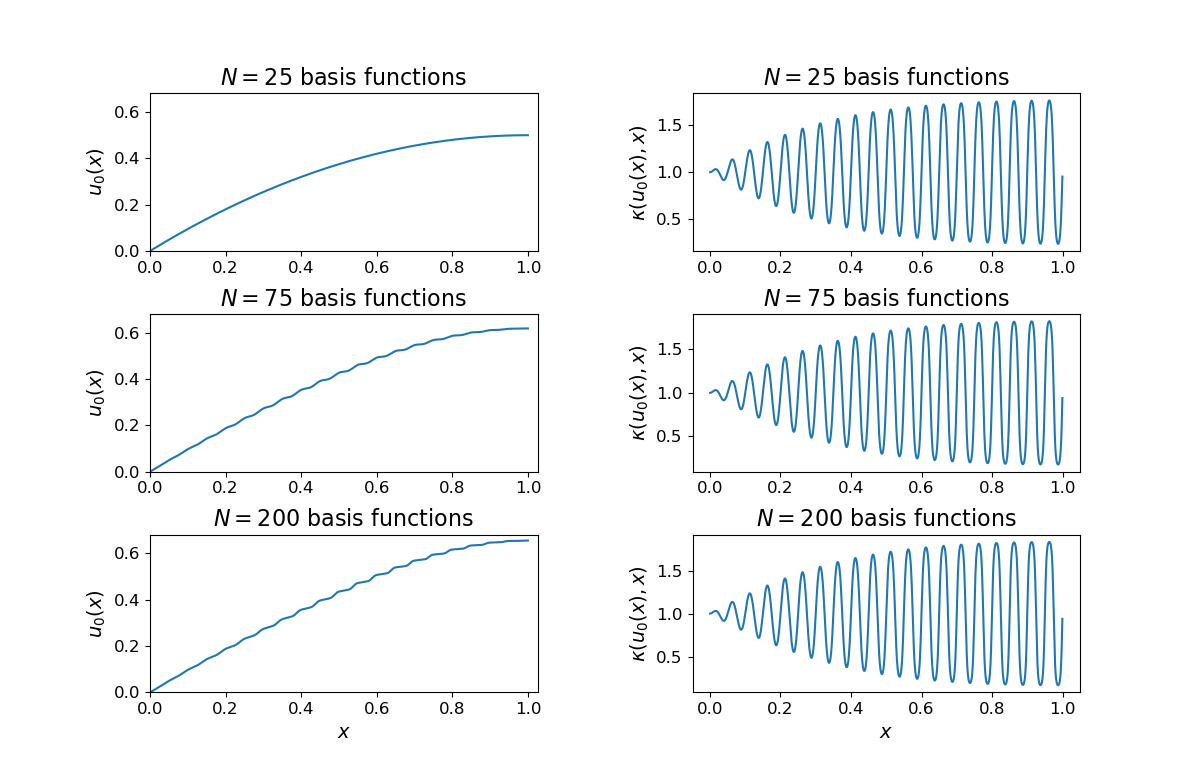}
\caption{Sweeping over different resolutions of the basis expansion of the multiscale temperature, we see that coarse bases cannot capture even the macroscopic component of the solution.}
\label{basis_sweep}
\end{figure}

\noindent which enforces both boundary conditions automatically. By solving the nonlinear system of equations corresponding to the Galerkin weak form of the governing equation, we solve the multiscale problem with $N=25,75,200$ basis functions in the discretization at $\eta=0.05$. Figure \ref{basis_sweep} shows the computed temperature fields, as well as the material property at the converged temperature distribution. Only when the basis is capable of resolving the high-frequency fluctuations is the low-frequency part of the solution captured. We verify that further expanding the size of the basis beyond $N=200$ does not change the computed solution. Given the high-frequency fluctuations in the material, the fluctuations in the response are small, as suggested by homogenization theory. The system response can thus be approximated with a temperature-dependent conductivity with no small-scale fluctuations. We now turn to our parametric Deep Ritz approach to solve the cell problem and then compute the homogenized constitutive relation. We compare the temperature field computed from the homogenized constitutive relation to the fine discretization in Figure \ref{basis_sweep} as a benchmark.

\paragraph{} Per Eq. \eqref{homogenized}, the homogenized material property for the one-dimensional heat conduction problem and a unit cell domain is given by 
\begin{equation*}
    \hat \kappa(u_0) = \int_0^1 \kappa(u_0,y)\qty(1 + \pd{\chi_1(u_0,y)}{y}) dy,
\end{equation*}
\noindent where the unit corrector $\chi_1$ obeys the following parametric boundary value problem:
\begin{equation*}
    \pd{}{y}\qty( \kappa(u_0,y) \pd{\chi_1}{y}) + \pd{\kappa(u_0,y)}{y} = 0, \quad \chi_1(0) = \chi_1(1), \quad \pd{\chi_1(0)}{y}=\pd{\chi_1(1)}{y}, \quad \langle \chi \rangle = 0.
\end{equation*}

Discretizing the unit corrector as a function of both space and the  macroscopic temperature and setting the size of the cell domain to $L=1$, the following parametric Deep Ritz problem governs the corrector:
\begin{equation}\label{drm1}
    \Pi(\bs \theta) = \int_{u_{min}}^{u_{max}} \qty[  \int_0^1 \qty(\frac{1}{2} \kappa(u_0,y) \qty(1 +  \pd{\chi_1(u_0,y ; \bs\theta)}{y})^2  )dy + \lambda \qty( \int_0^1 \chi_1(u_0,y;\bs \theta) dy)^2]du_0, \quad \underset{\bs\theta}{\text{argmin }} \Pi(\bs \theta).
\end{equation}

\begin{figure}[hbt!]
\centering
\includegraphics[width=0.99\textwidth]{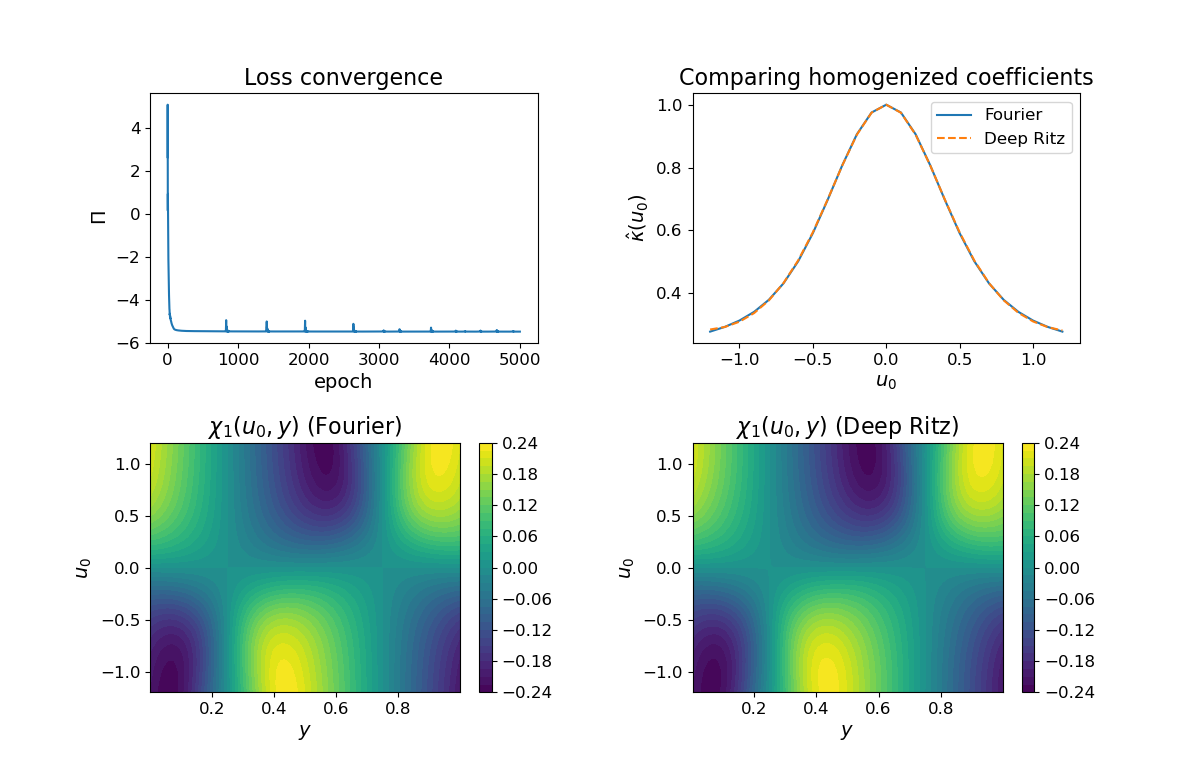}
\caption{Results for the one-dimensional Fourier-type heat conduction problem solved with the parametric Deep Ritz method. The energy converges in the allotted $5000$ epochs, and we compare the homogenized coefficients from the two numerical approximations of the parametric corrector field.}
\label{problem1}
\end{figure}

We use a two-hidden-layer multi-layer perceptron (MLP) network with width $50$ and hyperbolic tangent activations. The periodic boundary conditions on the corrector are embedded per Eq. \eqref{periodic_embedding}. The limits of integration for the macroscopic temperature are set at $u_{min}=-1.25$ and $u_{max}=1.25$ with $25$ evenly-spaced integration points. These limits set the range of validity of the parametric representation of the corrector, and must be chosen to reflect expected operating conditions of the macroscopic domain. We note that there is no reason to expect accurate solutions outside of the chosen range of macroscopic parameters, as the network is not trained to represent the corrector in these regions. The spatial integration grid is made up of $100$ evenly spaced points, and the penalty parameter is set at $\lambda=10^2$. The network is trained with Adam optimization for $5000$ epochs with a learning rate of $5 \times 10^{-3}$. We compare the solution obtained from Eq. \eqref{drm1} with a numerical solution obtained using the weak form of the cell problem and a periodic Fourier basis with the constant mode removed up to a frequency of $50\cdot 2\pi$, which provides sufficient resolution for the cell problem. With the Fourier basis, we carry out a ``brute force'' solution, where the cell problem is solved at each integration point of the macroscopic temperature grid. See Figure \ref{problem1} for the training convergence, a comparison of the computed homogenized coefficients from the two discretizations, and a visualization of the parametric unit corrector field. The parametric energy loss converges in the allotted $5000$ steps, and the homogenized coefficients closely match. To quantify the error between the two solutions, we look at the relative $L_1$ error between the two corrector fields given by 

\begin{equation*}
    \mathcal E = \frac{\int_{u_{min}}^{u_{max}} \int_0^1 | \chi_1^{\text{DRM}} - \chi_1^{\text{Fourier}}| dy du_0}{\int_{u_{min}}^{u_{max}} \int_0^1 | \chi_1^{\text{Fourier}}| dy du_0}.
\end{equation*}

\begin{figure}[hbt!]
\centering
\includegraphics[width=0.85\textwidth]{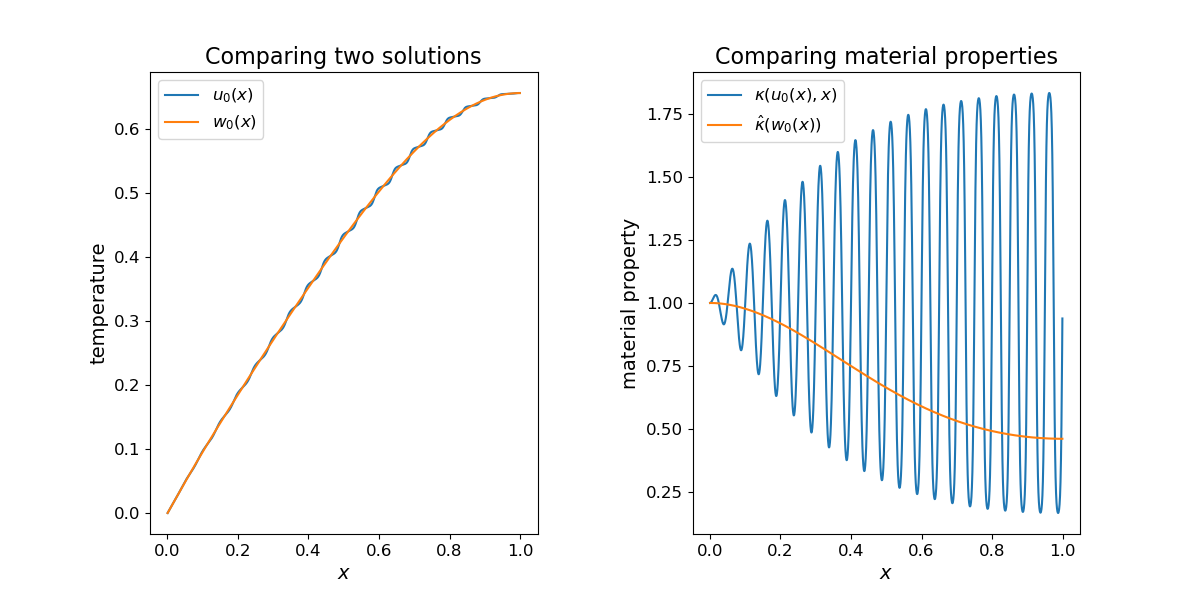}
\caption{The homogenized response accurately captures the low-frequency part of the multiscale solution, which dominates the response. We use $20\times$ fewer basis functions in computing the homogenized response. The homogenized material property only depends on the temperature, and thus exhibits no rapid fluctuations.}
\label{problem1_comparison}
\end{figure}

The error we obtain with the chosen hyperparameters is $\mathcal E = 0.7\%$. We remark that the Fourier discretization of the corrector ensures that it is zero mean, so the small error with the Deep Ritz solution shows that the mean penalty is sufficient to accurately enforce this constraint. We now use the homogenized coefficient we have computed to approximate a solution to Eq. \eqref{1d_ms}. As a result of the homogenization procedure, the spatial fluctuations of the temperature-dependent conductivity have been removed. We use the same spectral discretization to enforce the boundary conditions and discretize the temperature field, obtaining a numerical solution by forming and solving the weak form system. Figure \ref{problem1_comparison} shows a comparison between the multiscale solution of Eq. \eqref{1d_ms} and the homogenized solution, as well as a comparison of the multiscale and homogenized material properties. Note that $200$ basis functions are required to obtain the multiscale temperature field, whereas only $10$ are required for the homogenized temperature.

\subsection{Two-dimensional Fourier-type heat conduction}

\paragraph{} In this example and the following one, we focus exclusively on solving the cell problem (e.g., not on using the homogenized properties to compute the macroscopic response), verifying our parametric Deep Ritz solutions against traditional numerical methods. Here, we consider a two-dimensional Fourier-type heat conduction problem, where the cell problem represents a two-phase composite. The conductivity of each phase of the composite has a different temperature dependence. To define the inclusion phase of the composite, we introduce the following level-set field:
\begin{equation*}
    \phi(y_1,y_2) = 0.05 - (y_1-1/2)^2 - (y_2-1/2)^2 +1.5|(y_1-1/2)(y_2-1/2)|, \quad \mbf y \in [0,1]^2,
\end{equation*}
\noindent where $\phi>0$ is taken to represent the inside of the inclusion, $\phi<0$ outside, and $\phi=0$ represents the boundary between the two phases. Next, we introduce a smoothed step function $h(x)= ( 1 + \tanh(500x))/2$. With these in hand, we define the conductivity in the cell as 
\begin{equation}\label{problem2_conductivity}
    \kappa(u,y_1,y_2) = h(\phi(y_1,y_2)) + \frac{1}{2}(1 + u^2)( 1 - h (\phi(y_1,y_2))).
\end{equation}

\begin{figure}[hbt!]
\centering
\includegraphics[width=0.99\textwidth]{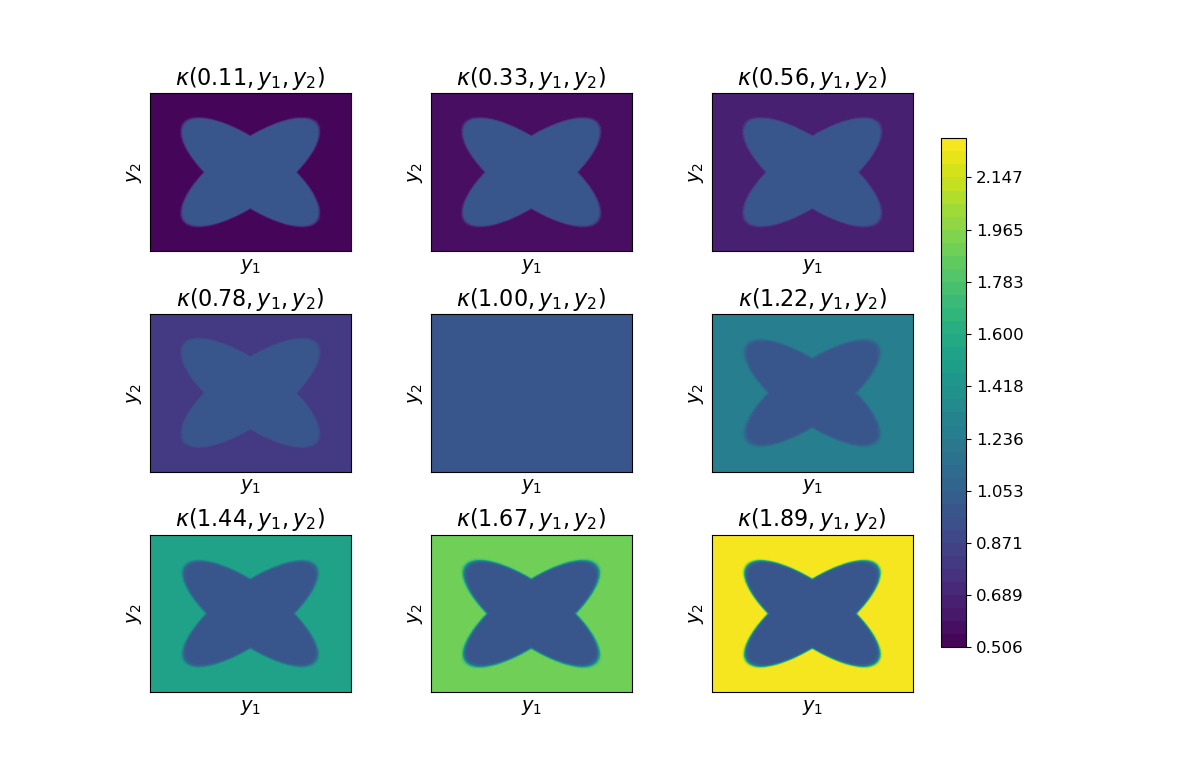}
\caption{The two-phase composite microstructure is built up of a petal-shaped inclusion and a surrounding matrix material. The conductivity of both phases depends on the temperature in different ways, and the cell has uniform conductivity at $|u|=1$. }
\label{problem2_conductivities}
\end{figure}

\begin{figure}[hbt!]
\centering
\includegraphics[width=0.99\textwidth]{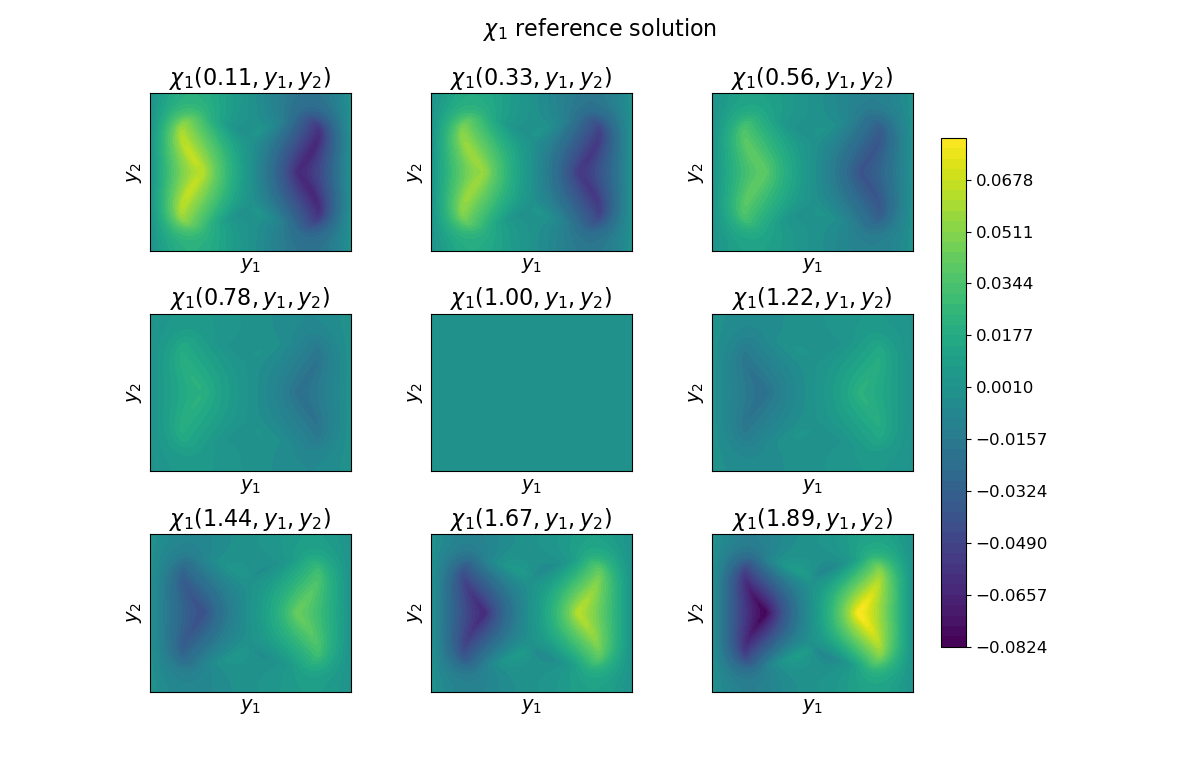}
\caption{We compute the corrector corresponding to a unit $y_1$ temperature gradient at each macroscopic temperature on the integration grid. We see that the qualitative behavior of the corrector is similar at different macroscopic temperatures, but the sign flips when the more conductive phase swaps at $u=1$.}
\label{chi1_ref}
\end{figure}

\begin{figure}[hbt!]
\centering
\includegraphics[width=0.99\textwidth]{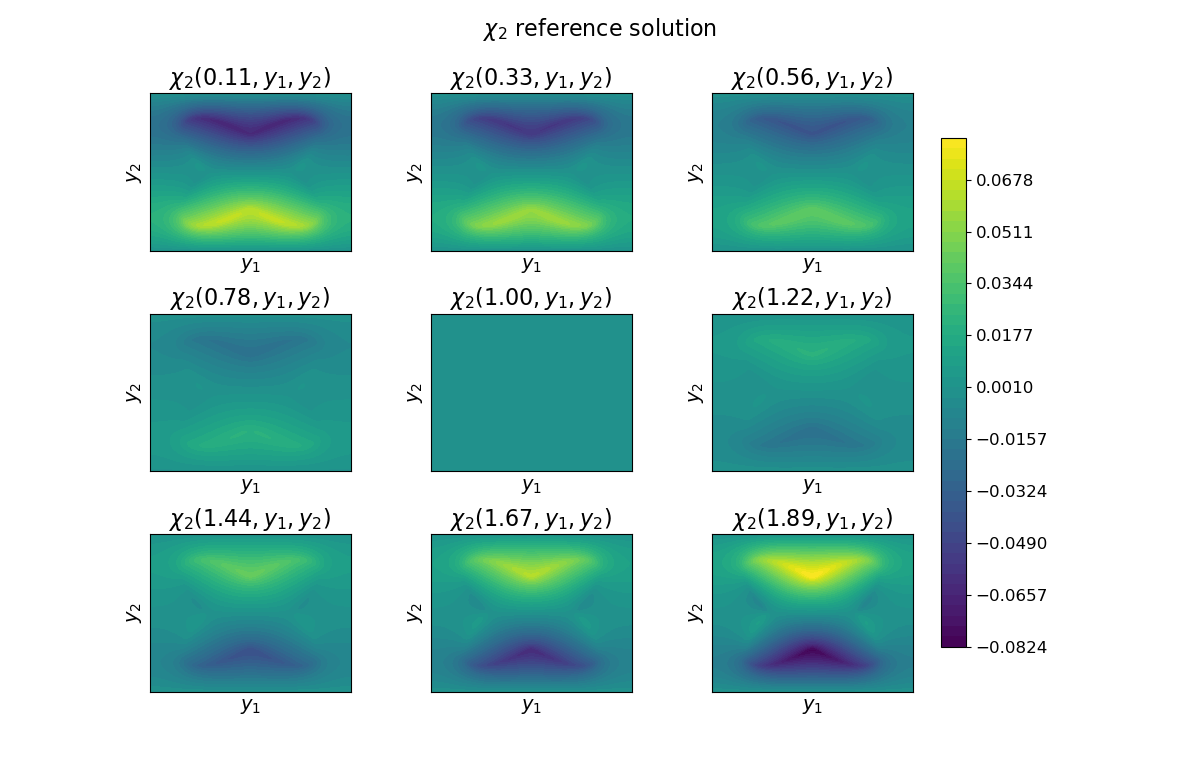}
\caption{The corrector corresponding to a unit $y_2$ temperature gradient is computed at each macroscopic temperature integration point. Note that when the corrector is computed in this brute force way, it is not represented as a continuous function of the macroscopic temperature. The unit corrector response to a vertical temperature gradient is a rigid rotation of the cell's response to a horizontal temperature gradient.}
\label{chi2_ref}
\end{figure}

See Figure \ref{problem2_conductivities} to visualize the cell conductivity over a range of macroscopic temperatures. In this problem, we take $u_{min}=0$ and $u_{max}=2$ and note that the conductivity is symmetric in the temperature about the origin; we only need to solve the cell problem at positive macroscopic temperatures. Because the constitutive law is linear in the macroscopic temperature gradient, the cell problems are linear, and thus we can solve for the corrector response to unit gradients in each coordinate direction. The parametric Deep Ritz problem is
\begin{equation*}
    \Pi( \bs \chi(u_0,\mbf y; \bs \theta)) = \int_{u_{min}}^{u_{max}} \qty[ \int  \frac{1}{2} \kappa(u_0,\mbf y)\qty( \delta_{ij} + \pd{\chi_i}{y_j} ) \qty( \delta_{ij} + \pd{\chi_i}{y_j} )d\mbf y + \lambda \Big \lVert \int \bs \chi d\mbf y \Big\rVert^2 ] du_0, \quad \underset{\bs \theta}{\text{argmin }} \Pi( \bs \chi(u_0,\mbf y ; \bs\theta) ),
\end{equation*}
\noindent where $\chi_i$ is the corrector corresponding to an applied unit temperature gradient in the $y_i$ direction and $\lambda$ is a penalty parameter that enforces that each component of the corrector vector is independently zero mean. Per Eq. \eqref{periodic_embedding}, we enforce the periodic boundary conditions by embedding the cell coordinate $\mbf y$. Like the previous example, the cell problem is parameterized only by the macroscopic temperature. To compute a reference solution, we use a two-dimensional periodic Fourier basis with the constant mode removed. To build this basis, we take a tensor product of the real representation of the Fourier basis given in terms of sines and cosines, then remove only the constant-constant tensor product mode. We take all modes up to a frequency of $25 \cdot 2\pi$ in the univariate Fourier basis, which corresponds to $51$ one-dimensional Fourier basis functions and $51^2-1=2600$ two-dimensional basis functions. We use a uniform $125 \times 125$ integration grid and midpoint quadrature to perform integrals. The reference solution for the unit corrector vector is computed with the Galerkin weak form of the cell problem of Eq. \eqref{cell_problem_unit}. Because this problem is parameterized by the macroscopic temperature, we carry out a brute force solution strategy, where the cell problem is solved at every integration point in the macroscopic temperature grid. Because the cell conductivity in Eq. \eqref{problem2_conductivity} is symmetric around the origin in the temperature, we take the integration grid for the macroscopic temperature to be $[0,2]$ with $9$ equally spaced integration points. See Figures \ref{chi1_ref} and \ref{chi2_ref} for the two components of the unit corrector vector at integration points on the macroscopic temperature grid. As expected, we see that the corrector components are zero when the cell conductivity is uniform at $u=1$. We use the numerical solution computed with the Fourier basis as a reference against which the Deep Ritz solution can be compared.

\begin{figure}[hbt!]
\centering
\includegraphics[width=0.99\textwidth]{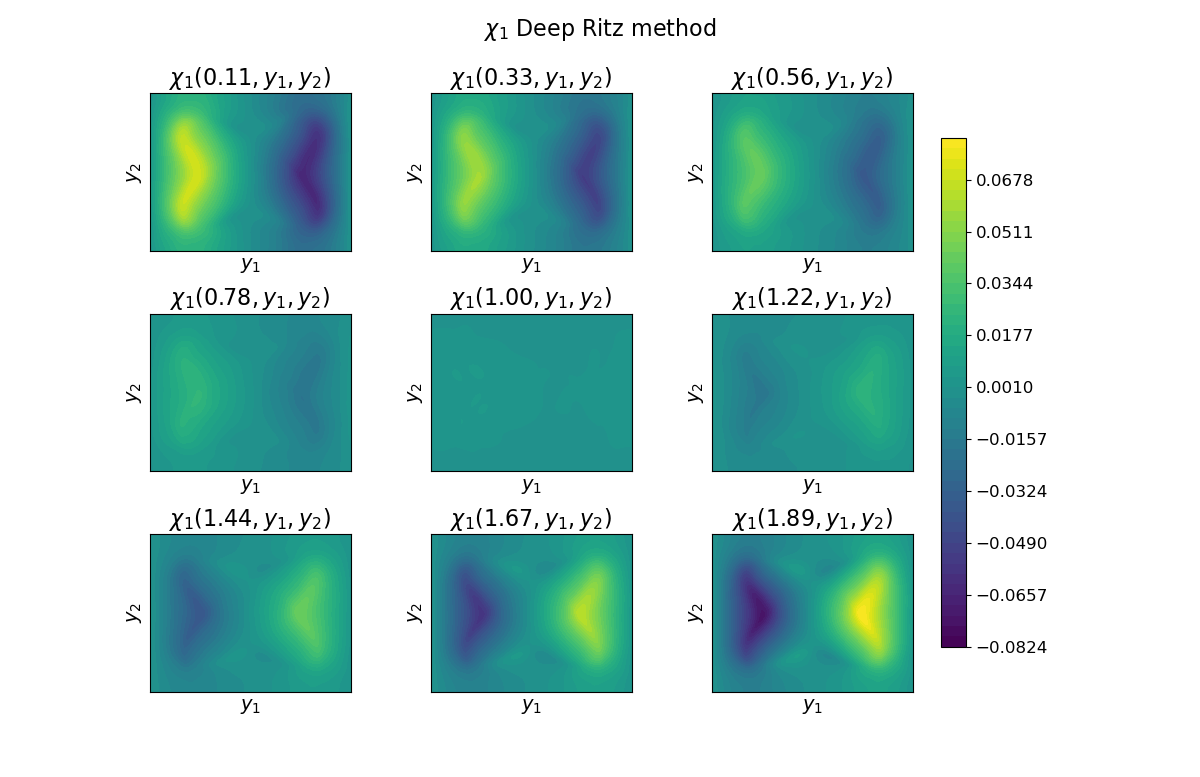}
\caption{The corrector response to a unit $y_1$ heat flux computed with the neural network discretization. While we show the corrector field on integration points of the macroscopic temperature grid, the neural network provides the corrector as a continuous function of the macroscopic temperature.}
\label{chi1_drm}
\end{figure}

\begin{figure}[hbt!]
\centering
\includegraphics[width=0.99\textwidth]{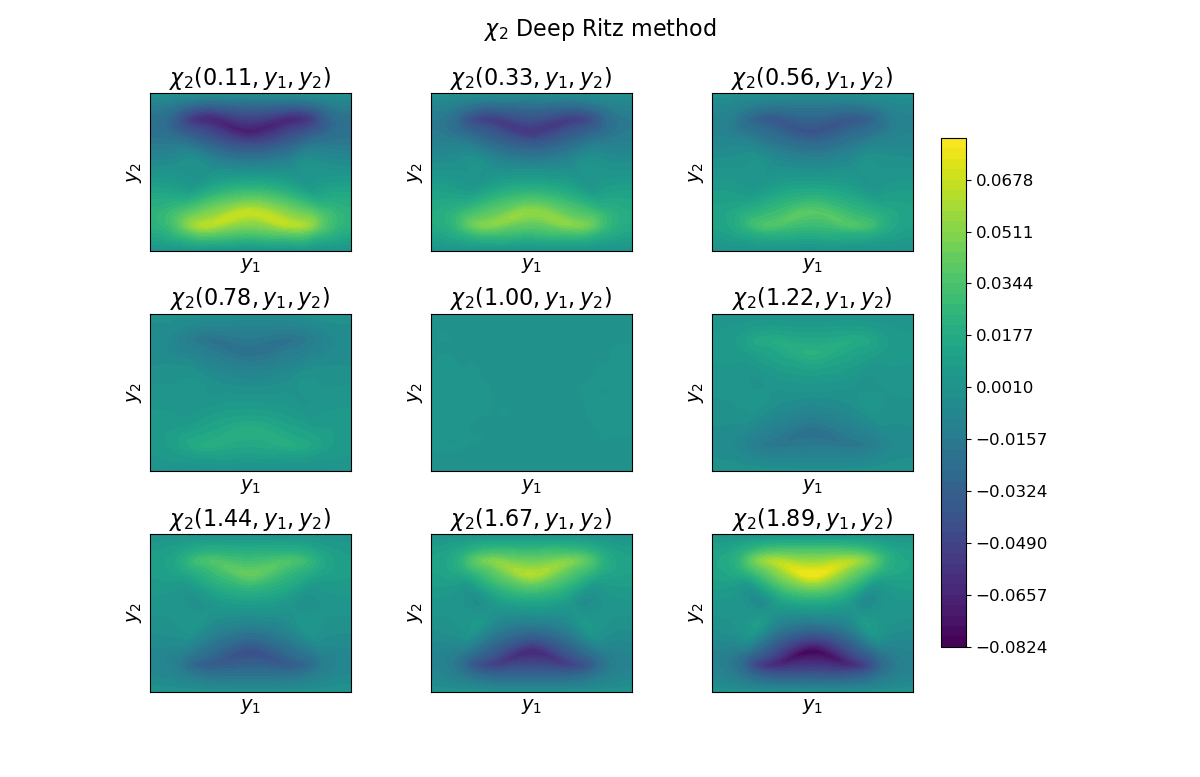}
\caption{The corrector response to a unit $y_2$ heat flux computed with the neural network discretization.}
\label{chi2_drm}
\end{figure}

\paragraph{} To solve the parametric Deep Ritz problem, the corrector field is represented using an MLP network with three hidden layers of width $50$. Because we do not expect the neural network to have high-frequency oscillatory components like the Fourier basis, we coarsen the integration grid to $60 \times 60$ uniformly spaced integration points with midpoint quadrature. We minimize the parametric Deep Ritz objective using Adam optimization for $10000$ epochs with a learning rate of $1 \times 10^{-3}$. The penalty parameter enforcing the zero mean condition on the corrector components is $\lambda= 100$. See Figures \ref{chi1_drm} and \ref{chi2_drm} for the parametric corrector field computed from the Deep Ritz method. Our error metric for the two fields is computed as follows:
\begin{equation}
    \mathcal E = \frac{1}{2} \qty( \frac{\int_{u_{min}}^{u_{max}}\int| \chi_1^{\text{DRM}} - \chi_1^{\text{Fourier}} | d \mbf y du_0}{\int_{u_{min}}^{u_{max}}\int| \chi_1^{\text{Fourier}} | d \mbf y du_0} + \frac{\int_{u_{min}}^{u_{max}}\int| \chi_2^{\text{DRM}} - \chi_2^{\text{Fourier}} | d \mbf y du_0}{\int_{u_{min}}^{u_{max}}\int| \chi_2^{\text{Fourier}} | d \mbf y du_0} ),
\end{equation}

\begin{figure}[hbt!]
\centering
\includegraphics[width=0.85\textwidth]{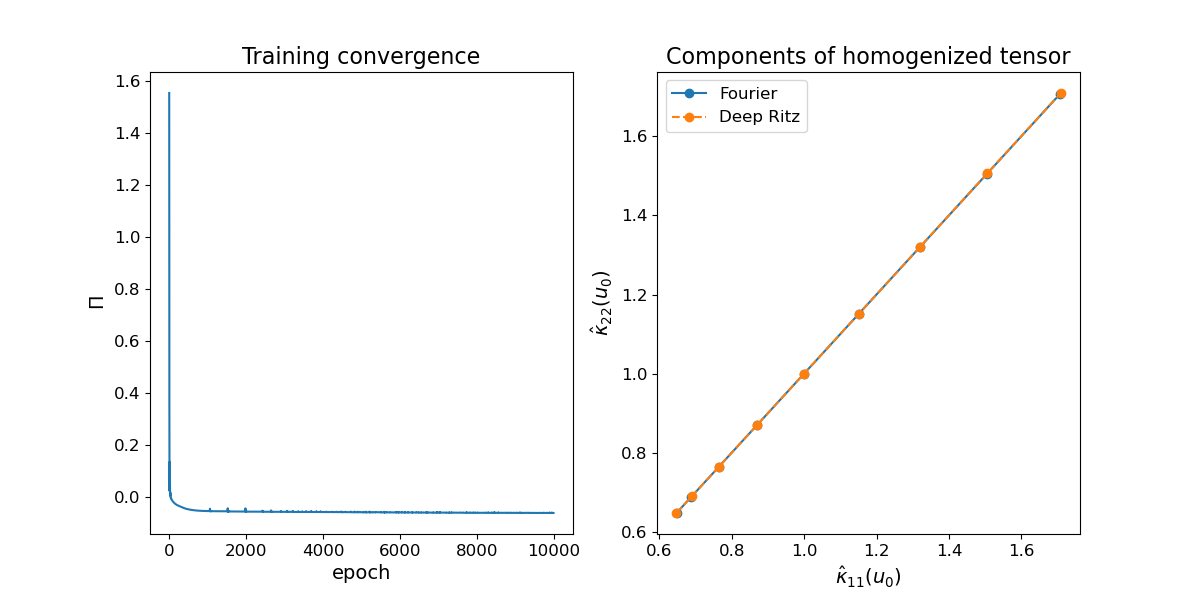}
\caption{Comparing the components of the homogenized conductivity tensor computed from both the Fourier and neural network discretizations of the solution. We visualize the homogenized tensor as a curve parameterized by the macroscopic temperature $u_0$. Because the off-diagonal term of the homogenized conductivity is negligibly small, the parameterized curve is plotted in the $\hat \kappa_{11} - \hat \kappa_{22}$ plane.}
\label{conductivity_curve}
\end{figure}

\noindent which is simply the relative $L_1$ error averaged over the two components of the unit corrector vector. In this case, we obtain an error of $\mathcal E = 4.7\%$. We take this error to be driven in part by the small response of the cell near $u=1$, where small fluctuations of the corrector field away from zero correspond to large relative errors. We note that this is a challenging problem for neural networks, as the network, which has smooth activation functions, must represent a discontinuous temperature gradient at the interface between the two material phases. We can better assess the practical importance of these errors in the corrector field by computing the homogenized constitutive tensor from the unit corrector responses obtained from the Fourier discretization and the parametric Deep Ritz method. To visualize the differences between the computed conductivity tensors, we treat the three independent components of the symmetric homogenized conductivity as a curve in $\mathbb R^3$ parameterized by the macroscopic temperature $u_0$. In other words, we can visualize the curve $[ \hat \kappa_{11}(u_0) , \hat \kappa_{22}(u_0), \hat \kappa_{12} (u_0) ]^T$ as computed by the two approaches to obtaining the unit corrector. However, the off-diagonal component of the homogenized conductivity tensor is zero over all macroscopic temperatures in the case of the Fourier discretization, and has a maximum value of $10^{-4}$ in the case of the Deep Ritz discretization. Thus, we visualize only the curve defined by $[ \hat \kappa_{11}(u_0) , \hat \kappa_{22}(u_0) ]^T$. Recalling that the homogenized conductivity tensor is computed per Eq. \eqref{homogenized}, Figure \ref{conductivity_curve} shows the training convergence of the Deep Ritz solution and the components of the homogenized tensor computed from both discretizations as a function of the macroscopic temperature. The maximum error magnitude between any components of the homogenized conductivity over all macroscopic temperatures in the integration grid is $3\times 10^{-3}$, indicating the Deep Ritz solution to the cell problem leads to an accurate prediction of the homogenized response of the multiscale material.

\paragraph{} To showcase the benefit of the neural network discretization of the cell response, we can compute the homogenized constitutive tensor at a macroscopic temperature that does not lie on the integration grid. Because we have a continuous representation of the unit correctors, we can easily compute homogenized properties at any temperature in the range $[u_{min},u_{max}]$. We pick $u=0.2$, which lies roughly halfway between the two integration points at $u=0.11$ and $u=0.33$, and compare the predicted homogenized tensor from the neural network to a (new) solution obtained with the Fourier basis:
\begin{equation*}
    \bs{\hat \kappa}^{\text{Fourier}} = \begin{bmatrix}
        0.659 & 0 \\ 0 & 0.659
    \end{bmatrix}, \quad \bs{\hat \kappa}^{\text{DRM}} = \begin{bmatrix}
        0.660 & 0 \\ 0 & 0.660
    \end{bmatrix}.
\end{equation*}

Though the error between the corrector fields was $4.7\%$, the error in components of the homogenized tensor at a macroscopic temperature off the integration grid is only $0.15\%$. This suggests that it is not necessary to exactly resolve the corrector components in order to obtain high-accuracy estimates of the macroscopic material response. However, it is beneficial to know the corrector field in addition to the homogenized material properties, as this can be used to better understand the response of the multiscale body. For example, the microscale heterogeneity may create stress concentrations that represent important design considerations \cite{torisaki_shape_2023}. As such, we take another relevant metric of the model's performance to be its prediction of ``extreme events'' in the cell. In our heat transfer problem, we take this to be the maximum temperature in the cell. However, the maximum temperature is dependent on the temperature gradient, so we use the relative $L_1$ error of the maximum temperatures of the unit correctors as a proxy for this: 
\begin{equation*}
    \tilde {\mathcal E} = \frac{1}{2} \qty(  \frac{\int_{u_{min}}^{u_{max}} \Big|  \max(|\chi_1^{\text{DRM}}|) - \max(|\chi_1^{\text{Fourier}}|) \Big| du_0}{\int_{u_{min}}^{u_{max}} \max( |  \chi_1^{\text{Fourier}} |) du_0} + \frac{\int_{u_{min}}^{u_{max}} \Big|  \max(|\chi_2^{\text{DRM}}|) - \max(|\chi_2^{\text{Fourier}}|) \Big| du_0}{\int_{u_{min}}^{u_{max}} \max( |  \chi_2^{\text{Fourier}} |) du_0}  ),
\end{equation*}
\noindent where the $\max()$ is taken over the cell coordinate $\mbf y$. We take this to be a measure of the relative $L_1$ error of the maximum temperature predicted by the Deep Ritz method. The network performs more than twice as well on this metric, obtaining an error of $\tilde{\mathcal E} = 2\%$.

\paragraph{} As a final remark, we reiterate that the range of the macroscopic parameter---in this case, $u_0\in[0,2]$---is an important design consideration. If the parametric cell response is to be used in a macroscale simulation, this range reflects the analyst's prior knowledge about operating conditions. There is no reason to expect the trained network to accurately predict the cell response outside the chosen range. However, this parameter range can be chosen freely, and similar accuracy of the trained network is expected on and off the integration grid, so long as the density of parameter space integration points is approximately constant. For example, re-training the network with $u_0 \in [0,4]$ with $18$ parameter space integration points, we obtain similar errors in the parametric corrector ($\mathcal E = 3.1 \%$) and homogenized conductivity components.

\subsection{Two-dimensional $p$-Laplace heat conduction}

\paragraph{} In this example, we move away from Fourier heat conduction to the $p$-Laplace problem, where the cell problems are nonlinear, and thus we cannot build a homogenized conductivity tensor from the cell's response to unit temperature gradients. Instead, the cell problem is now parameterized by both the macroscopic temperature $u_0$ and the macroscopic temperature gradient $\nabla_x u_0$. We consider a cell problem designed to represent a two-dimensional metamaterial. Like the previous example, the cell comprises two materials with temperature-dependent conductivities. Instead of an inclusion in a matrix, the cell is a building block of a lattice-type structure. We can combine level sets for different bars to build this lattice structure:
\begin{equation*}
    \begin{aligned}
        \phi_1(y_1,y_2)= 0.025 - 10( y_1 - 1/2)^2, \\
        \phi_2(y_1,y_2)= 0.025 - 10( y_2 - 1/2)^2, \\
        \phi_3(y_1,y_2)= 0.025 - 5( y_1 + y_2 - 1)^2, \\
        \phi_4(y_1,y_2)= 0.025 - 5( y_1 - y_2)^2, \\
        \phi(y_1,y_2) = \max( \phi_1 , \phi_2 , \phi_3 , \phi_4 ).
    \end{aligned}
\end{equation*}

Using the same smoothed step function $h(\cdot)$ introduced previously, the conductivity of the cell is given by 
\begin{equation*}
    \kappa(u_0,y_1,y_2) = ( 1 + \exp(u_0/2) )h(\phi(y_1,y_2)) + \log(1+\exp(u_0))(1-h(\phi(y_1,y_2))).
\end{equation*}

\begin{figure}[hbt!]
\centering
\includegraphics[width=0.99\textwidth]{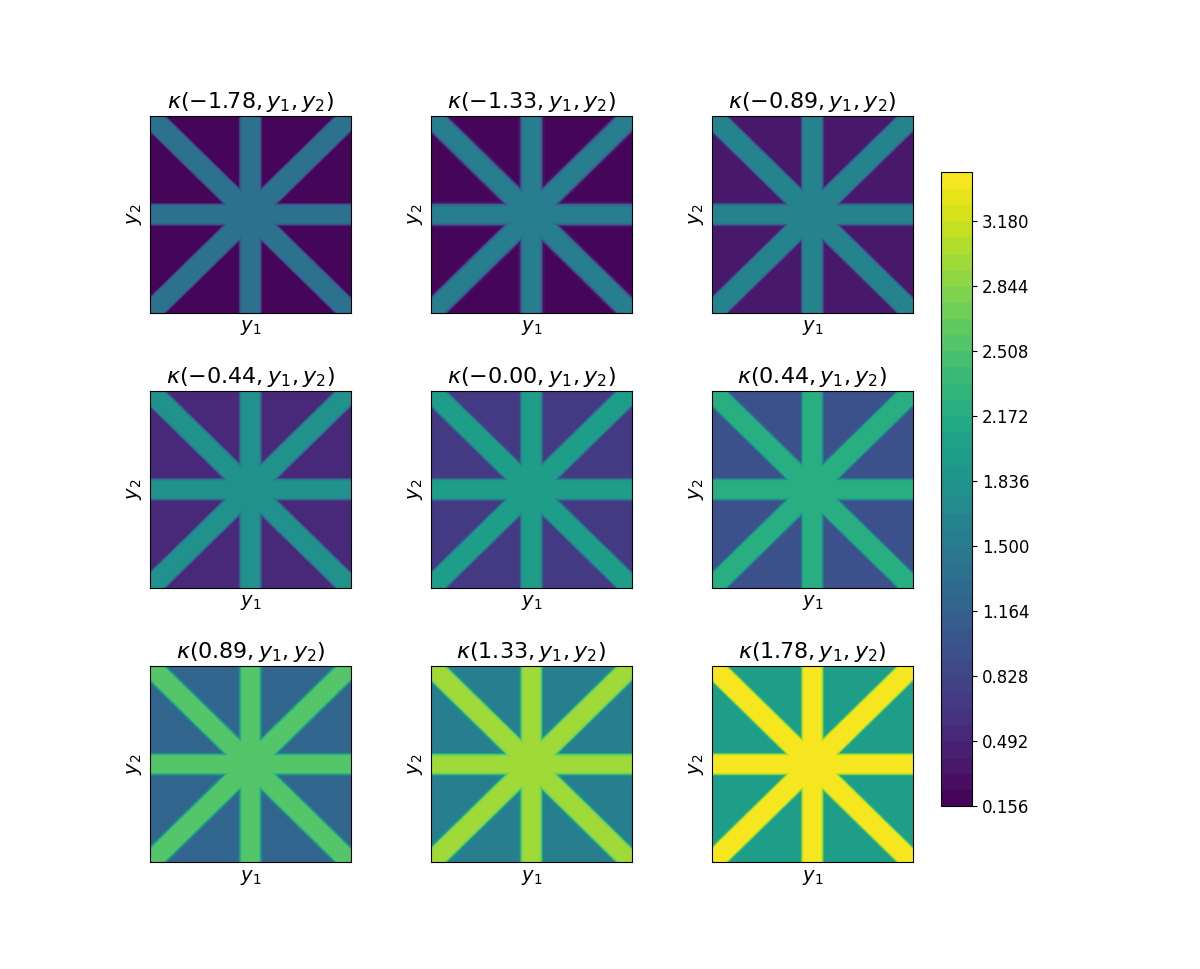}
\caption{The metamaterial cell problem, made up of two material phases with different temperature dependence. Here, the conductivity of the cell is shown over a range of macroscopic temperatures.}
\label{problem3_conductivities}
\end{figure}

See Figure \ref{problem3_conductivities} to visualize the cell conductivity over a range of temperatures. With this in hand, we write down the parametric Deep Ritz problem for $p$-Laplace cell problems with the help of Eq. \eqref{varp}:
\begin{equation}\label{plaplace_penalty}
    \Pi(  \chi( u_0 , \nabla_x u_0 , \mbf y; \bs \theta)) = \int \int\qty[ \int  \frac{1}{p} \kappa(u_0, \mbf y) \Big | \nabla_x u_0 + \nabla_y \chi \Big|^p d \mbf y + \lambda \qty(\int \chi d\mbf y)^2 ] du_0d(\nabla_x u_0 ), \quad \underset{\bs\theta}{\text{argmin }} \Pi( \bs \theta).
\end{equation}

Given that $\nabla_x u_0 \in \mathbb R^2$, the periodic embedding of Eq. \eqref{periodic_embedding}, and the lack of linearity to exploit, the corrector is now a function $\chi: \mathbb R^7 \rightarrow \mathbb R$. Note that with two spatial coordinates and three parameters, the dimension of the problem is $5$, but the periodic embedding introduces two additional input dimensions. To generate reference solutions at given macroscopic temperatures and temperature gradients, we can use the two-dimensional Fourier basis with the constant mode removed and the weak form of the $p$-Laplace cell problem, which is obtained from stationarity of the variational energy:
\begin{equation}\label{pstationarity}
    \delta \qty(\int \frac{1}{p} \kappa(u_0 ,\mbf y)\Big| \nabla_x u_0 + \nabla_y \chi \Big|^p d\mbf y ) = \int \kappa(u_0, \mbf y) | \nabla_x u_0 + \nabla_y \chi |^{p-2}(\nabla_x u_0 + \nabla_y \chi) \cdot \delta \nabla_y \chi d\mbf y = 0,
\end{equation}

\begin{figure}[hbt!]
\centering
\includegraphics[width=0.99\textwidth]{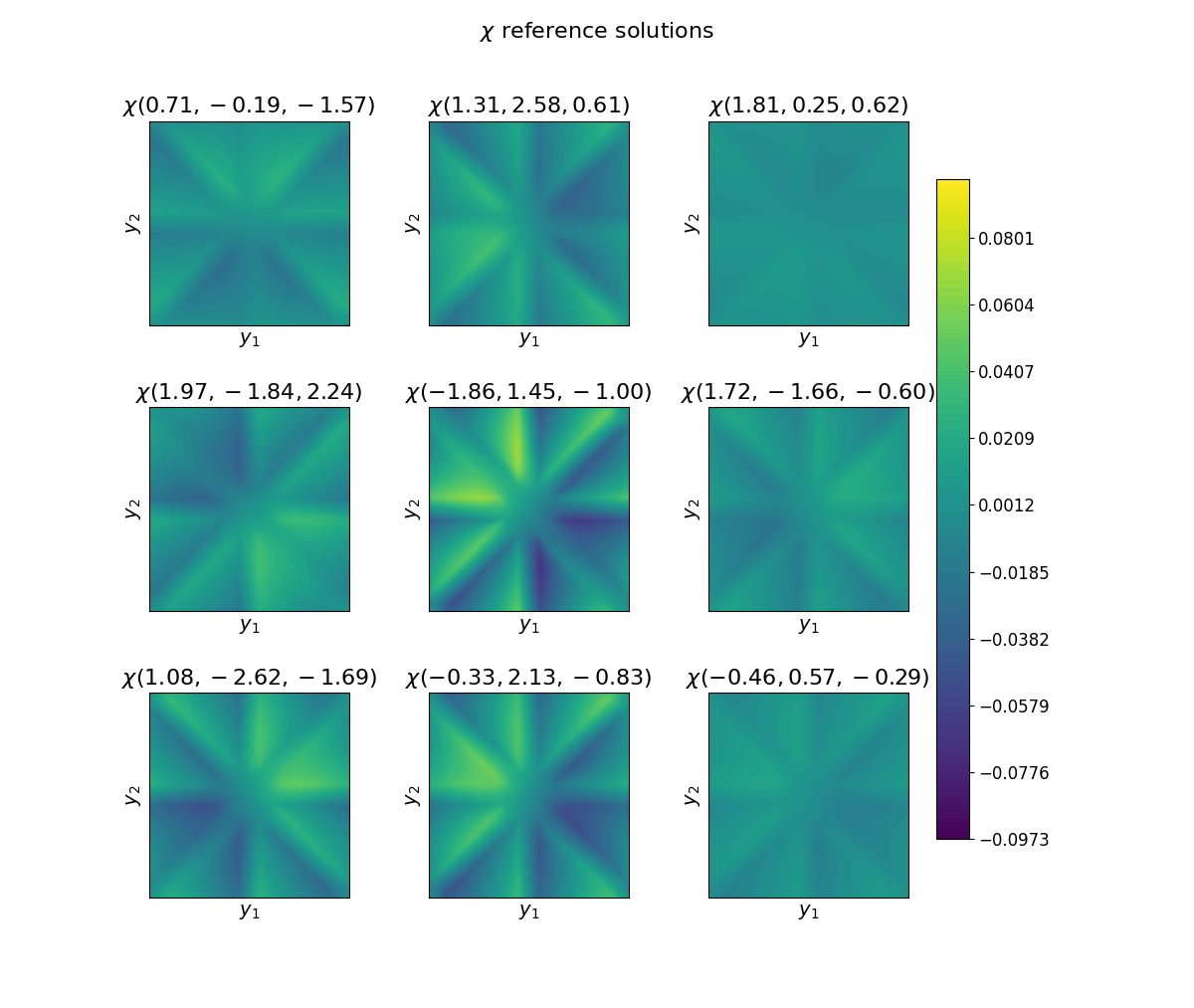}
\caption{Reference solutions obtained from the Fourier discretization of the $p$-Laplace cell problem. Because we cannot exploit the linearity of the cell problem, it is necessary to solve for the corrector at all settings of the macroscopic temperature and temperature gradient. We show the computed corrector $\chi(u_0,\nabla_x u_0)$ at $9$ randomly sampled parameter settings.}
\label{problem3_ref}
\end{figure}

\noindent where both $\chi$ and its variation $\delta \chi$ are discretized with the zero-mean two-dimensional Fourier basis. We use an integration grid of $125 \times 125$ uniformly spaced points and take frequencies of the one-dimensional Fourier basis functions up to $30 \cdot 2\pi$, corresponding to $3720$ basis functions. Note that the resulting system of equations is now nonlinear, so we solve it with Newton's method by computing the Hessian of the variational energy analytically. See Figure \ref{problem3_ref} for reference solutions of the cell problem at randomly sampled macroscopic temperatures and temperature gradients with $p=4$. In previous examples, we have constructed an integration grid for both physical and parameter space. We now have two spatial variables, two components of the macroscopic temperature gradient, and a macroscopic temperature on which the corrector field depends. If we use $75$ spatial integration points in each dimension, and $10$ integration points for each parameter dimension, this is an integration grid of $75^2 \times 10^3=5.6 \times 10^6$ integration points. Thus, if we train the network for $10^4$ iterations, building a full integration grid for the parametric Deep Ritz objective corresponds to an intractable number of neural network and gradient evaluations over the course of training, especially as the parameter grid is further refined. To alleviate this issue, we opt to use Monte Carlo integration for the integrals over the parameters, while keeping the spatial integration on a fixed grid. We uniformly sample points $u_0 \in [-2,2]$ and $\nabla_x u_0 \in [-3,3]^2$, which defines the region in parameter space over which the parameterized corrector is valid. The stochastic version of the parametric Deep Ritz objective is thus
\begin{equation*}
    \Pi(  \chi( u_0 , \nabla_x u_0 , \mbf y; \bs \theta)) \approx \frac{1}{B}\sum_{i=1}^B\qty[ \int  \frac{1}{p} \kappa(u_0^i, \mbf y) \Big | \nabla_x u_0^i + \nabla_y \chi (u_0^i , \nabla_x u_0^i,  \mbf y ; \bs \theta)\Big|^p d \mbf y + \lambda \qty(\int \chi(u_0^i,\nabla_x u_0^i,\mbf y; \bs \theta) d\mbf y)^2 ] ,
\end{equation*}

\begin{figure}[hbt!]
\centering
\includegraphics[width=0.99\textwidth]{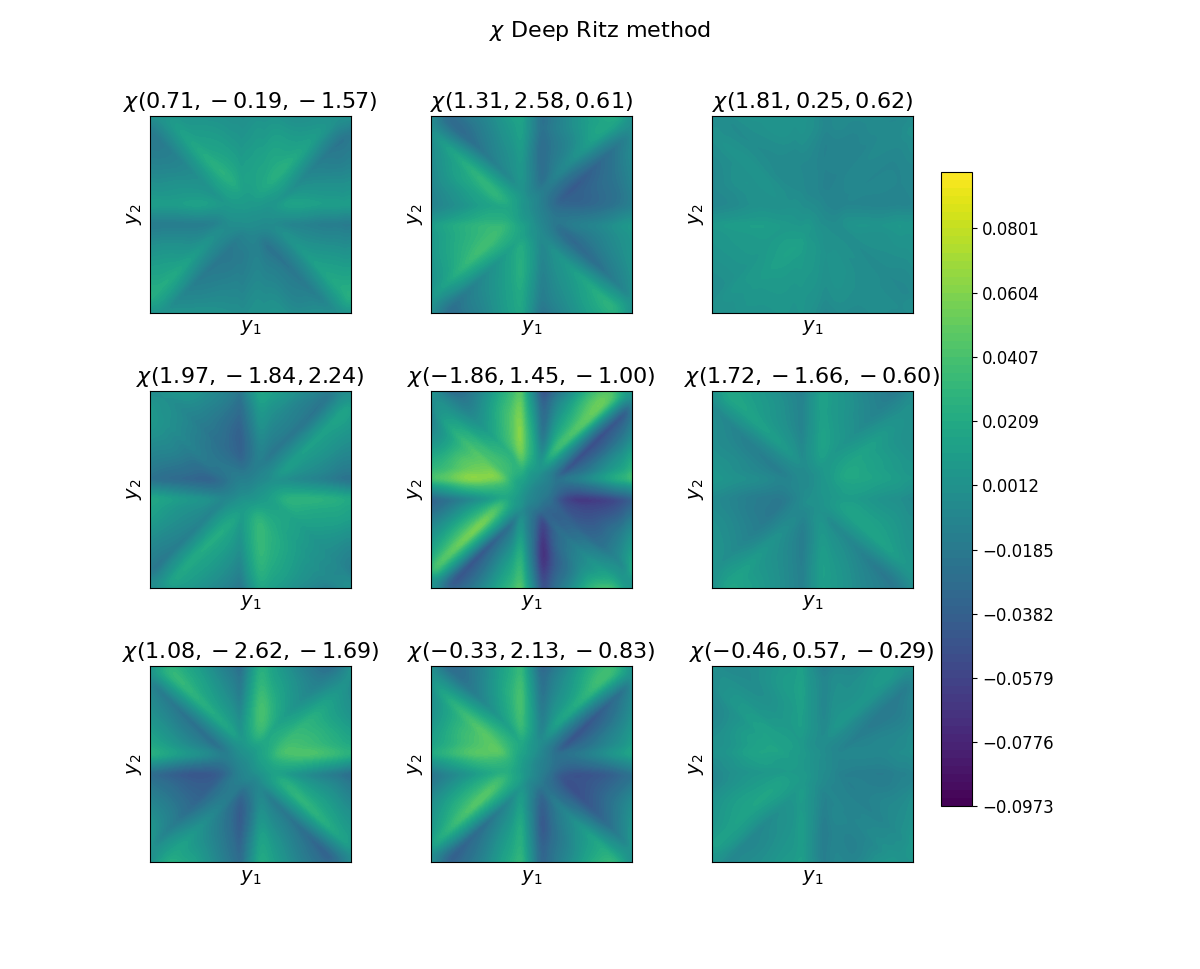}
\caption{Solutions obtained from the neural network at the same random settings of the macroscopic parameters as the reference solutions shown in Figure \ref{problem3_ref}.}
\label{problem3_nn}
\end{figure}

\noindent where $B$ is the size of the Monte Carlo batch and $u_i \overset{\text{i.i.d.}}{\sim} \mathcal U([-2,2])$ and $\nabla_x u_i \overset{\text{i.i.d.}}{\sim} \mathcal U([-3,3]^2)$. We use a three-layer network of width $75$, a batch size of $B=25$, and train the network for $10000$ epochs with Adam optimization at a learning rate of $10^{-3}$. See Figure \ref{problem3_nn} to visualize the corrector fields obtained at the same random parameter settings as the reference solution. The corrector fields from the parametric Deep Ritz method qualitatively match the reference corrector fields, and also obtain the correct order of magnitude of the temperature. However, in some cases, the cell temperature around the bar structure is not as sharply defined as the reference solutions. The relative $L_1$ error between the two fields is given by
\begin{equation*}
    \mathcal E = \frac{\int \int \int  \int | \chi^{\text{DRM}} - \chi^{\text{Fourier}}| d\mbf y d u_0 d(\nabla_x u_0)}{\int \int \int  \int | \chi^{\text{Fourier}}| d\mbf y d u_0 d(\nabla_x u_0)},
\end{equation*}
\noindent where we approximate the parameter integrals with Monte Carlo integration. We expect errors when using Monte Carlo integration for the Deep Ritz objective, as the parameterized variational energy is only approximated at every optimization epoch. The relative $L_1$ error we obtain with $50$ random samples of the macroscopic parameters is $\mathcal E=17 \%$, which is rather large. In many problems of practical interest, the macroscopic flux computed from the cell problem is more important than exactly matching the cell temperature. In addition to this point, we observe that the corrector does not make a significant contribution to the cell temperature, as its maximum value is generally small compared to the imposed macroscopic temperature. Thus, we compute the macroscopic fluxes from the reference solutions and Deep Ritz representation and compare these, taking this to be the relevant representation of the accuracy of the computed solution. Using Eq. \eqref{pflux}, we use the corrector obtained from each solution method to compute the error between the two fluxes as
\begin{equation}\label{p_flux_error}
    \tilde{\mathcal E} = \frac{\int \int \mbf \lVert \mbf q^{\text{DRM}} - \mbf q^{\text{Fourier}} \rVert du_0 d(\nabla_x u_0)}{\int \int \mbf \lVert  \mbf q^{\text{Fourier}} \rVert du_0 d(\nabla_x u_0)}.
\end{equation}

Using $50$ random samples from the uniform distribution over macroscopic parameters, we obtain a relative error in the computed macroscopic fluxes of $\mathcal E = 0.2\%$. This example shows that even if there are errors in the corrector field, the homogenized response of the cell can still be accurately computed, which is driven by gradients of the corrector rather than the corrector itself. This is fortunate, as precisely estimating the correctors is a challenging, high-dimensional nonlinear problem. Using Monte Carlo integration streamlines the solution process by avoiding forming a dense $5$-dimensional integration grid over both space and the macroscopic parameters, but it comes at the cost of some accuracy of the computed correctors. The homogenization problem is particularly forgiving here, as these errors are washed out in computing the macroscopic response, as the small relative $L_1$ error of the homogenized flux shows. 

\paragraph{} We remark that the high accuracy of the fluxes computed from the parameterized corrector suggests that this approach is a useful way to generate training data for a data-driven approach to homogenization. If one were to take an approach like \cite{dana_machine_2021}, where the homogenized material response is learned as a function of the macroscopic parameters from data, it is necessary to generate a large amount of input-output pairs for the neural network surrogate. With our method, once the network is trained, it is possible to sample continuously from the parameter space to generate training data, without recomputing cell problems. We take this to be a benefit of the neural network parametrization and the Monte Carlo integration---it is possible to accurately learn the continuous dependence of the cell solution on the parameters without ever forming a dense integration grid in the parameter space. This is an asset even if the goal is to generate training pairs for a data-driven homogenized constitutive model.

\section{Comparing the cost of repeated cell solutions}

\paragraph{} In this section, we compare the computational cost of our method in repeatedly computing the cell response to that of an $\text{FE}^2$-type approach and a non-parametric neural network discretization. With our method, the cell response is computed as a continuous and differentiable function of the macroscopic parameters in an offline fashion, so no online PDE solves are necessary. In contrast, when using a traditional (e.g., non-parametric) discretization of the cell problem, it is necessary to recompute the corrector at each setting of the macroscopic parameters. Using two-dimensional $p$-Laplace heat conduction for the cell problem, we compare our method against a non-parametric spectral discretization in by timing the calculation of the macroscopic flux and its parameter gradient at $100$ random settings of the macroscopic temperature and temperature gradient, which we use as a surrogate for the repeated evaluations of the cell problem encountered during Newton iterations in an $\text{FE}^2$-type multiscale simulation. Then, we compute the cost of a single Deep Ritz solution with the non-parametric neural network discretization given in \cite{li_energy-based_2026}, and use this to determine the number of cell solutions at which the parametric solve becomes advantageous. These example problems are meant to assess the increase in computational efficiency of multiscale simulations with a pre-computed parametric representation of the cell response.

\paragraph{} In the following two examples, we are interested in the cost of repeated evaluations of the macroscopic response of a $p$-Laplace cell problem. The temperature-dependent material property of the cell is defined by 
\begin{equation*}
    \begin{aligned}
        a(y_1,y_2,u_0) = (y_1-1/2)\cos(u_0) + (y_2-1/2) \sin(u_0) ,\\
        b(y_1,y_2,u_0) = - (y_1-1/2) \sin(u_0) + (y_2-1/2) \cos(u_0), \\
        \phi(y_1,y_2,u_0) = 0.2 - 10 a(y_1,y_2,u_0)^2 - b(y_1,y_2,u_0)^2, \\
        \kappa(u_0,y_1,y_2) = 2h(\phi(y_1,y_2,u_0)) + 1-h(\phi(y_1,y_2,u_0)).
    \end{aligned}
\end{equation*}

\begin{figure}[hbt!]
\centering
\includegraphics[width=0.99\textwidth]{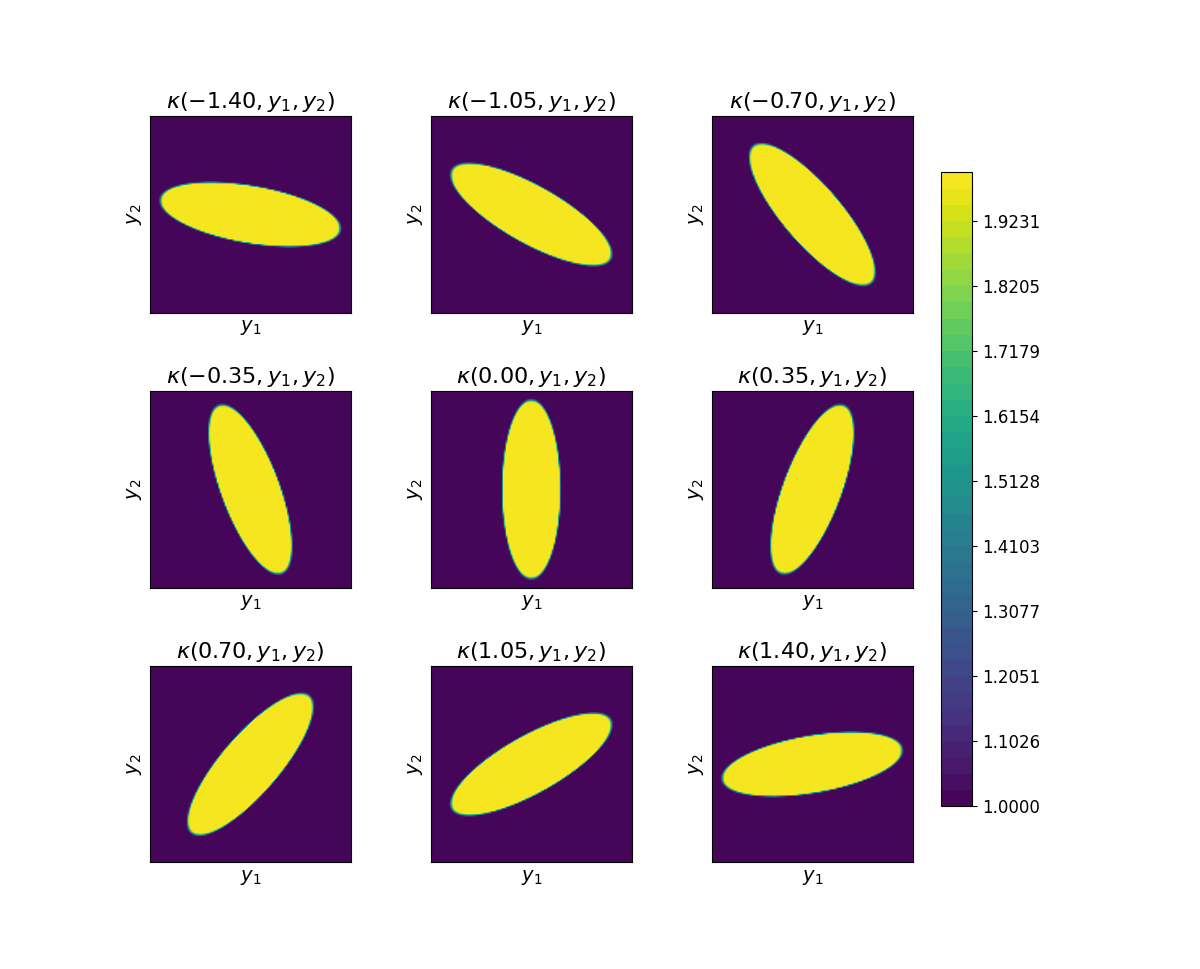}
\caption{The geometry of the inclusion in the cell changes with the macroscopic temperature. In particular, an elliptical inclusion rotates with the macroscopic temperature $u_0$. An interesting consequence of this choice of nonlinearity is that the anisotropy of the material depends on the macroscopic temperature. }
\label{rotating_ellipse}
\end{figure}

The conductivity of the cell is inspired by a phase change problem, where the geometry of the inclusion changes with the macroscopic temperature. However, this problem could also be conceptualized as solving the homogenized response of a parametric microstructure, where the rotation angle of the inclusion represents uncertainty about the geometry of the cell problem. This demonstrates that our method is indifferent to whether the parametric dependence of the cell problem arises from nonlinear physics or cell geometry. See Figure \ref{rotating_ellipse} to visualize the dependence of the cell material property on the macroscopic temperature. With the given temperature-dependent conductivity, recall that in the $p$-Laplace setting, the parametric Deep Ritz problem for the corrector is given by Eq. \eqref{plaplace_penalty}:
\begin{equation*}
    \Pi(  \chi( u_0 , \nabla_x u_0 , \mbf y; \bs \theta)) = \int \int\qty[ \int  \frac{1}{p} \kappa(u_0, \mbf y) \Big | \nabla_x u_0 + \nabla_y \chi \Big|^p d \mbf y + \lambda \qty(\int \chi d\mbf y)^2 ] du_0d(\nabla_x u_0 ), \quad \underset{\bs\theta}{\text{argmin }} \Pi( \bs \theta).
\end{equation*}

We now compare the computational cost of the solution obtained from our proposed method against existing strategies in the literature, namely 1) a traditional spectral discretization, which is used to compute the macroscopic flux and its gradient with respect to the macroscopic parameters, and 2) a non-parametric Deep Ritz method, where the corrector needs to be re-computed at each setting of the macroscopic parameters. 

\subsection{Comparison against $\text{FE}^2$ iterations}

\paragraph{} To solve this problem with a traditional spectral discretization, a stationary point of this energy functional is obtained per Eq. \eqref{pstationarity} at a given macroscopic temperature and temperature gradient. The corrector and its variation are then discretized in a fixed basis, and the resulting nonlinear system of equations is solved with Newton's method. Regardless of how the corrector is obtained, the macroscopic flux for a unit cell is given by 
\begin{equation*}
    \mbf q( u_0, \nabla_x u_0) = - \int \kappa(u_0 , \mbf y)| \nabla_x u_0 + \nabla_y \chi|^{p-2}( \nabla_x u_0 + \nabla_y \chi ) d\mbf y.
\end{equation*}

\begin{figure}[hbt!]
\centering
\includegraphics[width=0.99\textwidth]{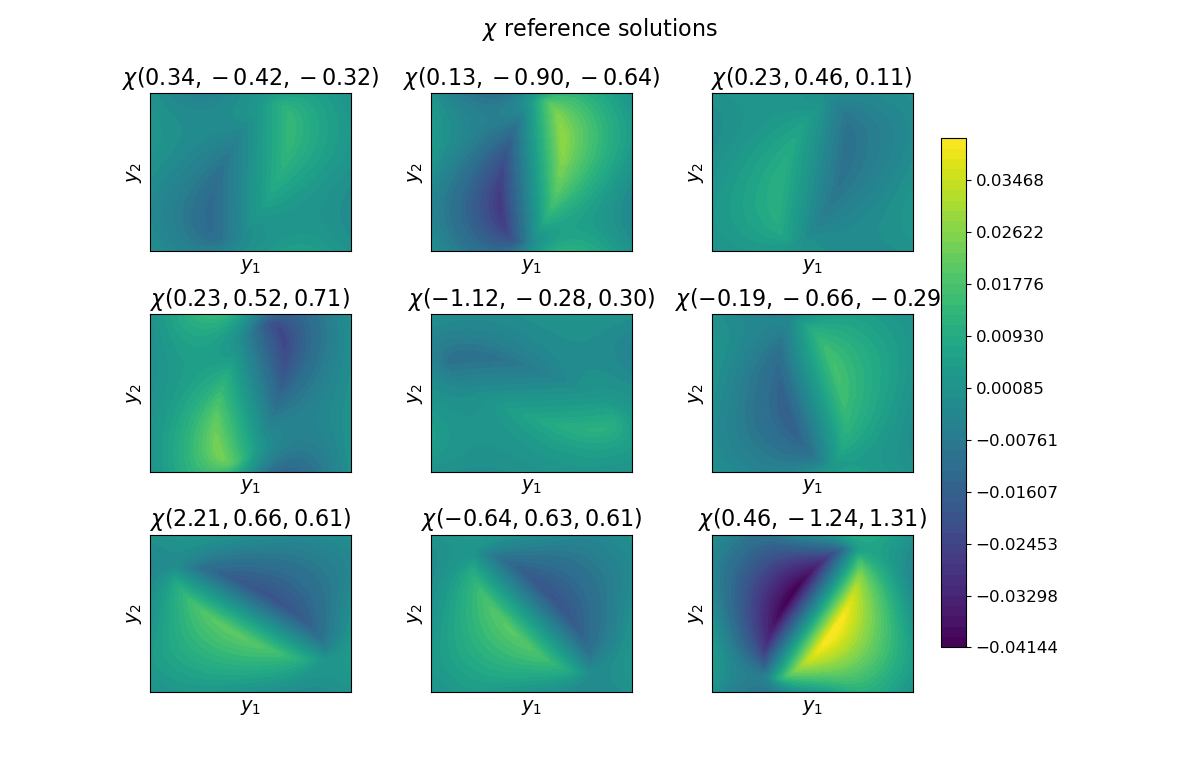}
\caption{Visualizing corrector fields $\chi(u_0, \nabla_x u_0)$ at randomly sampled macroscopic parameters. Recall that the elliptical inclusion rotates with the macroscopic temperature.}
\label{fe2_ref}
\end{figure}

When performing Newton iterations in a nonlinear multiscale simulation, it is necessary to compute the gradient of the macroscopic flux with respect to the macroscopic temperature and temperature gradient. When the corrector is obtained with the parametric Deep Ritz method, it depends continuously on the macroscopic parameters, and we can use automatic differentiation to compute $\partial \mbf q / \partial u_0$ and $\partial \mbf q / \partial(\nabla_x u_0)$. However, sensitivity analysis is required to compute the gradient of the flux when using a traditional discretization of the corrector field. Appendix D outlines how sensitivity derivatives of the macroscopic flux are computed by differentiating through the cell problem.

\paragraph{} To compare the cost of the parametric neural network discretization and the cell problem solved with a traditional discretization, we compute $\mbf q(u_0,\nabla_x u_0)$, $\partial \mbf q / \partial u_0$, and $\partial \mbf q / \partial (\nabla_x u_0)$ at $100$ random settings of the macroscopic parameters, where $u_0 \sim N(0,1)$ and $\nabla_x u_0 \sim N(\mbf 0 , \mbf I)$. Note that these are the quantities from the cell problem that are required for Newton iterations in an $\text{FE}^2$ scheme. In our implementation of the sensitivity derivatives, we again use a two-dimensional periodic Fourier basis with the constant mode removed. For the cell problem, the residual system is solved using Newton's method with the Jacobian matrix computed analytically. For a fair comparison, we use the same integration grid for both the neural network and Fourier discretization, which is a $75 \times 75$ uniform grid. With the Fourier basis, we retain univariate Fourier modes up to a frequency of $27 \times 2\pi$, which yields $N=3024$ basis functions. For the neural network discretization, we use a three-layer hidden MLP network of width $75$, and the network is trained using Adam optimization with a learning rate of $10^{-3}$ for $20000$ epochs and a Monte Carlo integration batch size of $B=25$. As before, the network is trained to represent the dependence of the corrector on both space and the macroscopic temperature parameters. We draw the same $100$ random samples of these macroscopic parameters with both methods. See Figure \ref{fe2_ref} to visualize corrector fields computed with the Fourier discretization for $9$ randomly sampled macroscopic parameters. Once the network is trained, the macroscopic flux is computed by integrating the corrector and gradients of the macroscopic flux with the neural network discretization are computed with automatic differentiation in PyTorch, as $\chi$ is a continuous function of the macroscopic parameters $u_0$ and $\nabla_x u_0$. Using Eq. \eqref{p_flux_error}, the error we obtain in the computed macroscopic fluxes is given by $\tilde {\mathcal E} = 0.4\%$. We also compute the error in the gradients computed from the parametric neural network discretization as
\begin{equation*}
    \frac{ \sum_{i=1}^{100} \sum_{j=1}^2 \sum_{k=1}^3 | \partial q^{\text{DRM}}_j(\mathcal M_i) / \partial \mathcal M_k - \partial q_j^{\text{Fourier}}(\mathcal M_i) / \partial \mathcal M_k|  }{ \sum_{i=1}^{100} \sum_{j=1}^2 \sum_{k=1}^3 | \partial q_j^{\text{Fourier}}(\mathcal M_i) / \partial \mathcal M_k|} = 0.8\%,
\end{equation*}
\noindent where $\mathcal M = [u_0 , \nabla_x u_0]^T$ is the collection of macroscopic parameters. Because a solution to a nonlinear system is not required at every macroscopic parameter setting, the $100$ macroscopic calculations are performed in $2.7$ seconds with the neural network discretization. In contrast, when the corrector needs to be found with online calculations, it takes $948.7$ seconds to perform these calculations with the Fourier discretization. Of course, implementations vary in efficiency---we acknowledge that some efficiency gains may be possible in our implementation of the Fourier-based homogenization and sensitivity analysis. That being said, our implementation is efficient in that only one nonlinear solve is required at every macroscopic parameter setting, and all gradients are computed analytically, thus avoiding expensive finite differencing operations. Rather than claiming that specific efficiency gains should be expected using the neural network discretization, the purpose of this experiment is simply to illustrate the potential speed-ups for nonlinear multiscale analysis that arise from a parametric representation of the corrector field. All computations are performed on an Apple M2 8-core processor (4 performance and 4 efficiency cores) with 8 GB unified memory, running Python 3.11.5 and PyTorch 2.1.2.

\subsection{Comparison against non-parametric Deep Ritz}

\paragraph{} In \cite{li_energy-based_2026}, a Deep Ritz method is proposed to solve the cell problems arising from nonlinear homogenization in the context of elasticity. However, the authors compute the homogenized response of the cell at given macroscopic parameters, meaning that the network needs to be retrained when these parameters change. In the case of $p$-Laplace heat transfer, a non-parametric formulation of the cell problem is 
\begin{equation}\label{nonparametric}
    \tilde \Pi(  \tilde \chi(\mbf y; \bs \theta) ; u_0 , \nabla_x u_0) = \ \int  \frac{1}{p} \kappa(u_0, \mbf y) \Big | \nabla_x u_0 + \nabla_y \tilde \chi \Big|^p d \mbf y + \lambda \qty(\int \tilde \chi d\mbf y)^2 , \quad \underset{\bs\theta}{\text{argmin }} \tilde \Pi( \bs \theta),
\end{equation}
\noindent where the modified energy objective $\tilde \Pi$ (rather than the solution) is now parameterized by $u_0$ and $\nabla_x u_0$. The notation $\tilde \chi$ is meant to indicate that, though this function still represents the cell response, it does not take in the macroscopic parameters as arguments. Of course, if the homogenized response of the cell at a single set of macroscopic parameters is of interest, it is cheaper to solve Eq. \eqref{nonparametric} as opposed to computing the full parametric response of the cell. However, because the network in Eq. \eqref{nonparametric} needs to be re-trained at each setting of the macroscopic parameters, we expect that the parametric representation becomes more efficient when repeated queries of the cell problem are required. To compute this ``cross-over'' point---the point at which the parametric representation becomes advantageous---we first obtain a parametric solution with Eq. \eqref{plaplace_penalty}. In particular, we use a three-layer MLP network with hyperbolic tangent activation functions and a width of $75$. The integration grid is $75 \times 75$ evenly spaced points in physical space, and a Monte Carlo integration batch size of $B=25$ for parameter space integration. We use Adam optimization with a learning rate $10^{-3}$ for $20000$ epochs. These are the same settings for training as in the previous problem, which already confirmed the accuracy of the converged solution. The training process takes $10588$ seconds to obtain the full parametric solution. In contrast, when obtaining a solution using Eq. \eqref{nonparametric} with a three hidden-layer network of width $75$, we find that $1500$ epochs are sufficient to obtain convergence using Adam optimization with a learning rate of $10^{-3}$. Over $5$ runs, the average time to compute a solution with randomly sampled macroscopic parameters is $27.3$ seconds. This means that after $10588/27.3\approx 388$ evaluations of the cell response, the parametric representation becomes more computationally efficient. Given that the cell response must be evaluated at every integration point in the macroscopic domain at every Newton iteration in an $\text{FE}^2$ scheme, it is reasonable to expect up to tens of thousands of cell evaluations in the process of obtaining a two-dimensional macroscopic solution. This suggests that a non-parametric Deep Ritz approach to approximating the cell response becomes infeasible when the response of the cell over a wide range of macroscopic parameters is required.


\section{Conclusion}

\paragraph{} In this work, we have introduced a method to solve the parametric cell problems that arise from nonlinear homogenization using the parametric Deep Ritz method. These cell problems are appealing for neural network discretizations because they are geometrically simple,  variational, and have boundary conditions that can be strongly enforced. Furthermore, the solution to the cell problem depends on one or more parameters of the macroscopic solution field and must be differentiated with respect to these macroscopic parameters, which makes a continuously differentiable neural network discretization in both physical and parameter space very natural. By comparing against a standard discretization strategy, we showed that the parametric neural network discretization accurately reproduces the reference solutions over a range of parameter settings, and that higher-dimensional parameter spaces can be efficiently handled with Monte Carlo integration. Then, we demonstrated the potential for expediting nonlinear multiscale simulations with the parametric representation of the corrector field, which achieved more than $300\times$ in our particular example problem. Of course, we note that this acceleration is only possible because the parametric representation of the corrector field is trained in an offline phase, similar to data-driven surrogate models for the cell response. However, unlike data-driven surrogates, our method does not require any training data, instead making use of the known physics of the cell problem. In fact, the method we propose could be a useful strategy for generating data for a surrogate model, as once it is trained, the parameter space can be sampled at any resolution, thus enabling efficient generation of large quantities of corrector fields. Another benefit of our method is that a new batch of Monte Carlo integration points can be used to randomly sample the parameter space at each optimization epoch. This is not the case with a data-driven surrogate, as the corrector field corresponding to each point in parameter space is required to supervise training; thus, integrating the parameter space must happen with a fixed set of samples. In contrast, the variational energy is straightforward to compute at any parameter setting, so better coverage of the high-dimensional parameter space is possible with the parametric Deep Ritz method and Monte Carlo integration.

\paragraph{} While we have studied the parametric Deep Ritz method for nonlinear homogenization in the context of heat transfer, future work may extend the method to the vector-valued problems of nonlinear elasticity. Subsequent studies should also focus on integrating this method into $\text{FE}^2$ schemes to better assess efficiency gains. Further research will be required to determine how many macroscopic parameters can be handled before the parameter integrals in our Deep Ritz formulation become intractably high-dimensional. We note that a benefit of the nonlinear homogenization problem is that pointwise knowledge of the corrector field is typically less important than integrals of its gradient. This means that in high-dimensional parameter spaces, it is possible that accurate macroscopic responses are computed even when the underlying corrector field is only approximate. In this sense, the nonlinear homogenization problem is a natural venue to push the limits of neural networks in handling high-dimensional inputs.

\section*{Acknowledgments}

This work was partially supported by the LEarning-Accelerated Domain Science (LEADS) SciDAC Institute and the National Laboratories of the Rockies. AD was also partially supported by the NASA Space Technology Research Institute Award ACCESS, under grant number 80NSSC21K1117.


\appendix

\counterwithin*{equation}{section}
\renewcommand\theequation{\thesection\arabic{equation}}

\section{Periodic boundary conditions}

\paragraph{} To understand the boundary conditions on $\chi$, we must first outline how the homogenized response of the cell, and thus the macroscopic constitutive relation, is obtained from the corrector. To find the cell response to an imposed gradient $\nabla_x u_0$ at temperature $u_0$, we compute the average flux through cross-sections of the cell. In particular, we compute the components of the macroscopic flux by averaging the normal components of heat flux from the cell temperature $u(\mbf y)$ over cross-sections of the cell.\footnote{The averaging operator avoids the macroscopic heat flux from being dependent on the choice of cell size $L$.} For example, the macroscopic flux in the $x_1$ direction is given by 
\begin{equation}\label{xsection}
    q_1( y_1, u_0,\nabla_x u_0) = \frac{1}{L}\int_0^L -\kappa(u_0,y_1,y_2) \pd{u}{y_1} dy_2 = \frac{1}{L} \int_0^L - \kappa(u_0,y_1,y_2)\qty( \pd{u_0}{x_1} + \pd{\chi}{y_1}(u_0 , \nabla_x u_0) )dy_2.
\end{equation}

\begin{figure}[hbt!]
\centering
\includegraphics[width=0.99\textwidth]{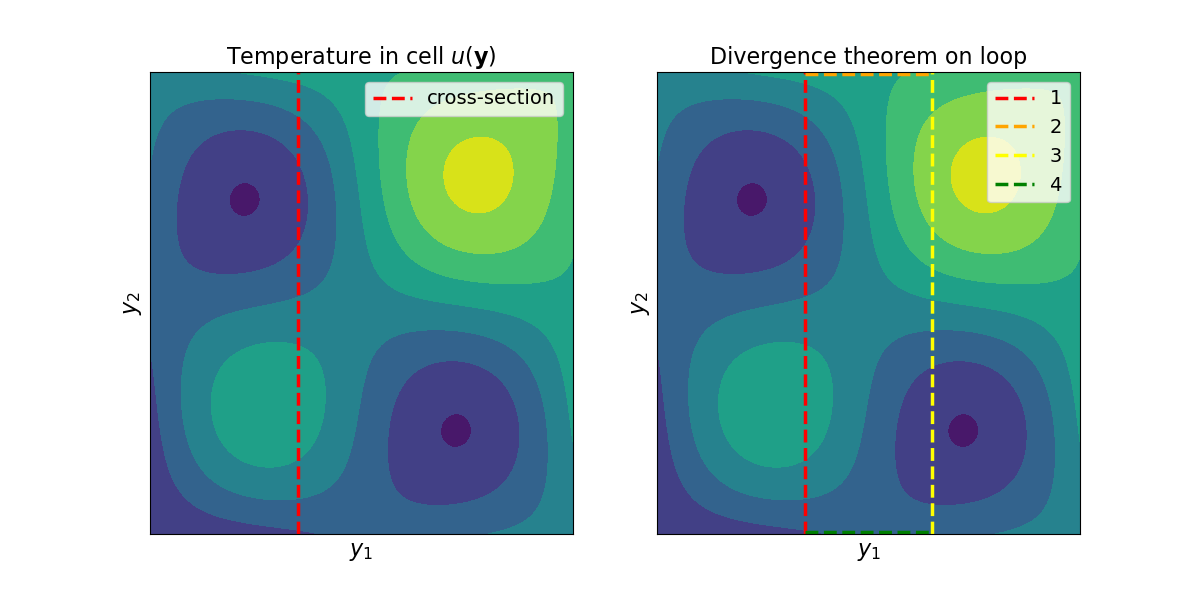}
\caption{We consider a two-dimensional nonlinear heat transfer problem to motivate periodic boundary conditions on the corrector field $ \chi(\mbf y)$. Because the cell problem states that the heat flux is divergence-free, the equality of integrated heat fluxes through different parallel cross-sections can be guaranteed by boundary properties of the corrector.}
\label{total_flux}
\end{figure}

Note that it is necessary to choose a position of the cross-section, here given by $y_1$. In order for there to be a well-defined macroscopic heat flux through the cell, we argue that the integrated heat flux should not depend on the position of the cross-section. In other words, the value computed from Eq. \eqref{xsection} should not depend on $y_1$, a condition which helps us determine the boundary conditions on $\chi$. Using the governing equation for the cell and the divergence theorem, we write
\begin{equation*}
    \int_{\mathcal V} \nabla \cdot \mbf q(\mbf y) d\mbf y = 0 = \int_{1+2+3+4} \mbf q(\mbf y) \cdot \mbf n dS,
\end{equation*}
\noindent where $\mbf n$ is the outward-facing normal vector, $\nabla \cdot \mbf q = 0$ expresses conservation of energy, the boundary segments in the line integral are shown in Figure \ref{total_flux}, and $\mathcal V$ is the volume enclosed by these segments. That there is zero net flux through the surface implies that the flux through the two vertical segments is equal when the boundary fluxes (horizontal segments) are equal. In other words,
\begin{equation*}
    \int_2 q_2 dy_1 = \int_4 q_2 dy_1 \implies \int_1 q_1 dy_2 = \int_3 q_1 dy_2 .
\end{equation*}

By inspecting the definition of the heat flux vector and recalling that the material property of the cell is periodic, it can be seen that the equality of the boundary fluxes is ensured when the corrector $\chi$ has either homogeneous Neumann or periodic boundary conditions on the upper and lower faces of the cell. As is standard in the homogenization literature, we take the corrector to be periodic in the $y_2$ direction. Similarly, the total flux in the vertical direction is given by 
\begin{equation*}
    q_2( y_2, u_0,\nabla_xu_0) = \frac{1}{L}\int_0^L -\kappa(u_0,y_1,y_2) \pd{u}{y_2} dy_1 = \frac{1}{L} \int_0^L - \kappa(u_0,y_1,y_2)\qty( \pd{u_0}{x_2} + \pd{\chi}{y_2}(u_0 , \nabla_x u_0) )dy_1.
\end{equation*}

Analogously, we require that the total vertical flux is independent of the height at which we take the cross-section. A similar divergence theorem argument shows that when the boundary fluxes on the vertical faces are equal, then so are the integrated fluxes at arbitrary $y_2$ cross-sections. This forces the corrector to be either homogeneous Neumann or periodic on the left and right faces of the cell, and we again choose periodicity. Thus, our argument that the macroscopic flux is computed by averaging the flux through cross-sections, and that the macroscopic flux is only well-defined when it is independent of cross-section, imposed particular boundary conditions on the corrector $\chi$. These arguments are all fundamentally tied to the periodicity of the material in the cell.

\section{Properties of homogenized constitutive relation}

\paragraph{} The first property of the temperature-dependent constitutive matrix that should be respected is symmetry. To prove symmetry, we test each governing equation for the corrector against the opposite corrector, integrate over the cell domain, and subtract one equation from the other:
\begin{equation*}
    \int \chi_2 \qty( \nabla_y \cdot\qty(\kappa(u_0,\mbf y)\qty(\nabla_y \chi_1 + \begin{bmatrix}
        1 \\ 0
    \end{bmatrix}))) d\mbf y -  \int \chi_1 \qty( \nabla_y \cdot\qty(\kappa(u_0,\mbf y)\qty(\nabla_y \chi_2 + \begin{bmatrix}
        0 \\ 1
    \end{bmatrix}))) d\mbf y = 0 .
\end{equation*}

Integrating the divergences by parts onto the test functions, using periodicity to cancel boundary terms, and eliminating a common term, this becomes
\begin{equation*}
    \int \qty( \kappa(u_0,\mbf y)\nabla_y \chi_2 \cdot  \begin{bmatrix}
        1 \\ 0 
    \end{bmatrix}  - \kappa(u_0,\mbf y)\nabla_y \chi_1  \cdot \begin{bmatrix}
        0 \\ 1 
    \end{bmatrix})d\mbf y = 0,
\end{equation*}
\noindent which proves the necessary equality for the symmetry of the effective conductivity for scalar cell conductivity fields, as the two off-diagonal terms of the homogenized conductivity are $\hat \kappa_{12} = \int \kappa \partial \chi _2 / \partial y_1 d\mbf y$ and $\hat \kappa_{21} = \int \kappa \partial \chi_1 / \partial y_2 d\mbf y$.

\paragraph{} The second law of thermodynamics puts additional restrictions on the homogenized conductivity tensor beyond symmetry. To see this, note that the entropy inequality for entropy density $s$ in a region $\Omega$ of a continuum body with no volumetric heat sources is
\begin{equation*}
    \int_{\Omega} \rho \dot s d\Omega \geq -\int_{\partial \Omega} \frac{\mbf q \cdot \mbf n}{u} dS = -\int_{\Omega} \nabla \cdot \qty(\frac{\mbf q}{u}) d\Omega,
\end{equation*}
\noindent where $\rho$ is the mass density, $\mbf q$ is the heat flux vector, $u$ is the temperature, and $\mbf n$ is the outward-facing normal vector to the boundary of the region $\Omega$. By localizing the integral, we obtain the point-wise statement of the entropy inequality:
\begin{equation*}  
    \rho \dot s \geq  -\nabla \cdot \qty(\frac{\mbf q}{u}) = -\frac{1}{u}(\nabla \cdot \mbf q) - \mbf q \cdot \nabla\qty(\frac{1}{u}).
\end{equation*}

The energy equation for the material point in the absence of source terms is $\rho \dot e = -\nabla \cdot \mbf q$, which we can substitute into the entropy inequality:
\begin{equation*}
    \rho \dot s \geq \frac{\rho \dot e}{u} + \frac{\mbf q \cdot \nabla u}{u^2}.
\end{equation*}

When there is no mechanical work done, e.g., the body is not undergoing deformations, the entropy change is purely energetic, meaning that $\dot e = u \dot s$. The entropy inequality then becomes
\begin{equation*}
    \frac{\mbf q \cdot \nabla u}{u^2} \leq 0,
\end{equation*}
\noindent which, given that the temperature is positive, shows that the heat flux must have a component opposite the temperature gradient to be physical. When the heat flux is given by a temperature-dependent Fourier's law, e.g., $\mbf q = - \bs \kappa(u) \nabla u$, this requires that the constitutive tensor $\bs \kappa(u)$ be positive semi-definite. Mathematically, this requires that $\nabla u^T \bs \kappa(u) \nabla u \geq 0$ for all temperatures $u$ and temperature gradients $\nabla u$. To prove positive semi-definiteness of the homogenized constitutive tensor, we define $F_i = \partial u / \partial x_i$, and show that
\begin{equation*}
    F_i F_j \int \kappa(u,\mbf y)\qty( \delta_{ij} + \pd{\chi_j}{y_i}) d\mbf y \geq 0 \quad \forall u,\mbf F.
\end{equation*}

To do this, we will show that the above quantity can be obtained through manipulations of another strictly positive quantity:
\begin{equation*}
\begin{aligned}
    \int \kappa(u,\mbf y) \qty(F_i + \pd{\chi_j}{y_i} F_j) \qty( F_i + \pd{\chi_k}{y_i} F_k) d\mbf y = \int \kappa(u, \mbf y)\qty( F_i F_i + F_i F_j\pd{\chi_j}{y_i}) d \mbf y + \int \kappa(u,\mbf y)\qty( F_i + \pd{\chi_j}{y_i} F_j) \pd{\chi_k}{y_i} F_k d \mbf y \\
    = F_i F_j \int \kappa(u,\mbf y) \qty(\delta_{ij} + \pd{\chi_j}{y_i}) d \mbf y - \int \pd{}{y_i} \qty( \kappa(u,\mbf y)\qty(F_i + \pd{\chi_j}{y_i} F_j)) \chi_k F_k d\mbf y + \int \kappa(u ,\mbf y) \qty(F_i + \pd{\chi_j}{y_i} F_j) \chi_k F_k n_i dS \\
    = F_i F_j \int \kappa(u,\mbf y) \qty(\delta_{ij} + \pd{\chi_j}{y_i}) d \mbf y,
\end{aligned}
\end{equation*}
\noindent where the boundary term cancels because of the periodic boundary conditions and the following term represents the weak form of the cell problem with a test function given by $\chi_k F_k$:
\begin{equation*}
    \int \pd{}{y_i} \qty( \kappa(u, \mbf y) \qty( F_i + \pd{\chi_j}{y_i} F_j)) \chi_k F_k d \mbf y = 0.
\end{equation*}

The above calculation shows that $\mbf F^T \hat{\bs \kappa}\mbf F $ is equal to a strictly positive quantity, assuming that the cell conductivity $\kappa(u,\mbf y)$ is also strictly positive. This proves the positive semi-definiteness of the homogenized conductivity tensor. We consider strong enforcement of these constitutive properties to be an advantage of elliptic homogenization, which requires passing through the cell problem. Purely data-driven methods, which fit the homogenized material response directly from data, do not, in general, guarantee that the learned constitutive relation is physical over the entire range of macroscopic parameters, unless specifically constrained to do so.

\section{Cell sensitivity analysis}

\paragraph{} Here, we outline the calculation of the gradient of the macroscopic flux---which requires differentiating through the cell problem---to compare the cost of this approach with the parametric neural network representation of the corrector. First, we compute the derivative of the macroscopic flux with respect to the macroscopic temperature:
\begin{equation}\label{grad1}
\begin{aligned}
    \pd{\mbf q}{u_0} = - \int \pd{\kappa(u_0 , \mbf y)}{u_0}| \nabla_x u_0 + \nabla_y \chi|^{p-2}( \nabla_x u_0 + \nabla_y \chi ) d\mbf y \\
    -\int \kappa(u_0,\mbf y) (p-2) |\nabla_x u_0 + \nabla_y \chi|^{p-4}(\nabla_x u_0 + \nabla_y \chi) \cdot \nabla_y\qty( \pd{\chi}{u_0} ) (\nabla_x u_0 + \nabla_y \chi)d\mbf y \\
    -\int \kappa(u_0,\mbf y) | \nabla_x u_0 + \nabla_y \chi |^{p-2} \nabla_y \qty(\pd{\chi}{u_0})d \mbf y.
\end{aligned}
\end{equation}

Next, the derivative of the flux with respect to the macroscopic temperature gradient is 
\begin{equation}\label{grad2}
\begin{aligned}
    \pd{ q_i}{\qty(\pd{u_0}{x_j})} = - \int \kappa(u_0,\mbf y) (p-2)|  \nabla_x u_0 + \nabla_y \chi |^{p-4}\qty(\pd{u_0}{x_k} + \pd{\chi}{y_k})\qty( \delta_{kj} + \pd{}{y_k} \pd{\chi}{\qty(\pd{u_0}{x_j})} )\qty(\pd{u_0}{x_i} + \pd{\chi}{y_i} ) d\mbf y \\
    -\int \kappa(u_0,\mbf y) | \nabla_x u_0 + \nabla_y \chi |^{p-2} \qty(\delta_{ij} + \pd{}{y_i} \pd{\chi}{\qty(\pd{u_0}{x_j})}) d\mbf y.
\end{aligned}
\end{equation}

In these expressions, we require the sensitivity derivatives of the corrector with respect to the macroscopic parameters. Per Eq. \eqref{pstationarity}, when discretizing the corrector in a basis per $\chi = \sum_{i=1}^N \chi_i h_i(\mbf y)$, the weak form system is 
\begin{equation}\label{res_sys}
     R_j(\bs \chi; u_0, \nabla_x u_0) = \int \kappa(u_0,\mbf y) \Big|\nabla_x u_0 + \sum_{i=1}^N \chi_i \nabla_y h_i (\mbf y)\Big|^{p-2}\qty(\nabla_x u_0 + \sum_{k=1}^N \chi_k \nabla_y h_k(\mbf y)) \cdot \nabla_y h_j(\mbf y)d\mbf y = 0.
\end{equation}

The sensitivity of the corrector coefficients with respect to the macroscopic temperature is computed with:
\begin{equation*}
\begin{aligned}
    \pd{}{u_0}  \mbf R( \bs \chi ; u_0 , \nabla_x u_0) = \mbf 0  = \pd{\mbf R}{\bs \chi} \pd{\bs \chi}{u_0} + \pd{\mbf R}{u_0} \implies \pd{\bs \chi}{u_0} = - \qty(\pd{\mbf R}{\bs \chi})^{-1}\qty(\pd{\mbf R}{u_0}),
\end{aligned}
\end{equation*}
\noindent where all quantities are evaluated at the corrector coefficients which solve the governing equation, e.g., $\mbf R(\bs \chi ; u_0, \nabla_x u_0) = \mbf 0$. This means that for every set of macroscopic parameters, a nonlinear system of equations needs to be solved. However, once the corrector coefficients are obtained through this solve operation, the sensitivity derivatives require only the solution to linear systems. Along similar lines, the sensitivity with respect to the macroscopic temperature gradient is 
\begin{equation*}
    \pd{\bs \chi}{(\nabla_x u_0)} = - \qty( \pd{\mbf R}{\bs \chi})^{-1} \qty(\pd{\mbf R}{(\nabla_x u_0)}).
\end{equation*}

The explicit derivatives of the residual system $\partial \mbf R / \partial u_0$ and $\partial \mbf R / \partial(\nabla_x u_0)$ are obtained analytically by differentiating Eq. \eqref{res_sys}. With the sensitivities of the corrector coefficients, the sensitivity fields can be constructed and integrated per Eqs. \eqref{grad1} and \eqref{grad2} to obtain the required gradients of the macroscopic heat flux. Like the previous $p$-Laplace example, we take $p=4$, which leads to some simplifications of the gradients of the macroscopic flux. With this choice for the nonlinearity and the basis representation of the corrector, the two gradients of the macroscopic flux with respect to the macroscopic parameters are:
\begin{equation*}
\begin{aligned}
    \pd{\mbf q}{u_0} = - \int \pd{\kappa(u_0 , \mbf y)}{u_0}| \nabla_x u_0 + \sum_i \chi_i \nabla_y h_i|^{2}( \nabla_x u_0 + \sum_i \chi_i \nabla_y h_i ) d\mbf y \\
    -\int 2\kappa(u_0,\mbf y) (\nabla_x u_0 + \sum_i \chi_i \nabla_y h_i) \cdot \qty(\sum_k \pd{\chi_k}{u_0} \nabla_y h_k  ) (\nabla_x u_0 + \sum_i \chi_i \nabla_y h_i )d\mbf y \\
    -\int \kappa(u_0,\mbf y) | \nabla_x u_0 + \sum_i \chi_i \nabla_y h_i |^{2}  \sum_i \pd{\chi_i}{u_0} \nabla_y h_i d \mbf y.
\end{aligned}
\end{equation*}

\begin{equation*}
\begin{aligned}
    \pd{ q_i}{\qty(\pd{u_0}{x_j})} = - \int 2\kappa(u_0,\mbf y) \qty(\pd{u_0}{x_k} + \sum_{\ell} \chi_{\ell} \pd{h_{\ell}}{y_k})\qty( \delta_{kj} + \sum_{\ell} \pd{\chi_{\ell}}{\qty(\pd{u_0}{x_j})} \pd{h_{\ell}}{y_k} )\qty(\pd{u_0}{x_i} + \sum_{\ell}\chi_{\ell} \pd{h_{\ell}}{y_i} ) d\mbf y \\
    -\int \kappa(u_0,\mbf y) | \nabla_x u_0 + \sum_{\ell} \chi_{\ell} \nabla_y h_{\ell} |^{2} \qty(\delta_{ij} + \sum_{\ell} \pd{\chi_{\ell}}{\qty(\pd{u_0}{x_j}) }\pd{h_{\ell}}{y_i}) d\mbf y.
\end{aligned}
\end{equation*}



\end{document}